\documentclass[journal]{IEEEtran}

\usepackage{cite}
\usepackage{fdsymbol}
\ifCLASSINFOpdf
  \usepackage[pdftex]{graphicx}
  \graphicspath{{../pdf/}{../jpeg/}}
  \DeclareGraphicsExtensions{.pdf,.jpeg,.png}
\else
  \usepackage[dvips]{graphicx}
  \graphicspath{{../eps/}}
  \DeclareGraphicsExtensions{.eps}
\fi
\usepackage{tikz}
\usetikzlibrary{positioning}
\usepackage{amsmath}
\usepackage{verbatim}
\usepackage[backref=section,
colorlinks=true,
citecolor=blue,
linkcolor=blue,
anchorcolor=red,
unicode=true,
urlcolor=blue]{hyperref}
\usepackage{bbding}
\usepackage{float}
\usepackage{multirow}
\usepackage{algorithmic}
\usepackage{color}
\usepackage{bm}

\usepackage{array}
\ifCLASSOPTIONcompsoc
  \usepackage[caption=false,font=normalsize,labelfont=sf,textfont=sf]{subfig}
\else
  \usepackage[caption=false,font=footnotesize]{subfig}
\fi

\ifCLASSOPTIONcaptionsoff
  \usepackage[nomarkers]{endfloat}
 \let\MYoriglatexcaption\caption
 \renewcommand{\caption}[2][\relax]{\MYoriglatexcaption[#2]{#2}}
\fi
\usepackage{url}
\usepackage{bbding}

\begin{document}
%
\title{Context-LGM: Leveraging Object-Context Relation for Context-Aware Object Recognition}
%
%

\author{Mingzhou~Liu,
        Xinwei~Sun\Envelope,
        Fandong~Zhang,
        Yizhou~Yu,
        and Yizhou~Wang

\thanks{\Envelope indicates corresponding authors.}
\thanks{Mingzhou Liu is with Computer Science Department, Peking University, 100871, Beijing, China (e-mail: liumingzhou@stu.pku.edu.cn).}
\thanks{Xinwei Sun is with Microsoft Research Asian, Beijing, 100080, China (e-mail: xinsun@microsoft.com).}
\thanks{Fandong Zhang is with DeepWise AI Lab, Beijing, 100080, China (zhangfdxk@163.com).}
\thanks{Yizhou Yu is also with DeepWise AI Lab, Beijing, 100080, China (yizhouy@acm.org).}
\thanks{Yizhou Wang is with Computer Science Department, Peking University, 100871, Beijing, China (e-mail: yizhou.wang@pku.edu.cn).}
}

%
%

\markboth{Journal of \LaTeX\ Class Files,~Vol.~XX, No.~X, September~2021}%
{M.Liu \MakeLowercase{\textit{et al.}}: Context-LGM: Leveraging Object-Context Relation for Context-Aware Object Recognition}
%



\maketitle

\begin{abstract}

Context, as referred to situational factors related to the object of interest, can help infer the object's states or properties in visual recognition. As such contextual features are too diverse (across instances) to be annotated, existing attempts simply exploit image labels as supervision to learn them, resulting in various contextual tricks, such as features pyramid, context attention, \emph{etc}. However, without carefully modeling the context's properties, especially its relation to the object, their estimated context can suffer from large inaccuracy. To amend this problem, we propose a novel \textbf{Context}ual \textbf{L}atent \textbf{G}enerative \textbf{M}odel (\textbf{Context-LGM}), which considers the object-context relation and models it in a hierarchical manner. Specifically, we firstly introduce a latent generative model with a pair of correlated latent variables to respectively model the object and context, and embed their correlation via the generative process. Then, to infer contextual features, we reformulate the objective function of Variational Auto-Encoder (VAE), where contextual features are learned as a posterior distribution conditioned on the object. Finally, to implement this contextual posterior, we introduce a Transformer that takes the object's information as a reference and locates correlated contextual factors. The effectiveness of our method is verified by state-of-the-art performance on two context-aware object recognition tasks, \textit{i.e.} lung cancer prediction and emotion recognition.\footnote{Codes will be made available after acceptance.}

\label{Sect_abstract}
\end{abstract}


%
\IEEEpeerreviewmaketitle

\section{Introduction}
\label{Sect_introduction}

\begin{figure}[!t]
\centering
\includegraphics[width=3.3in]{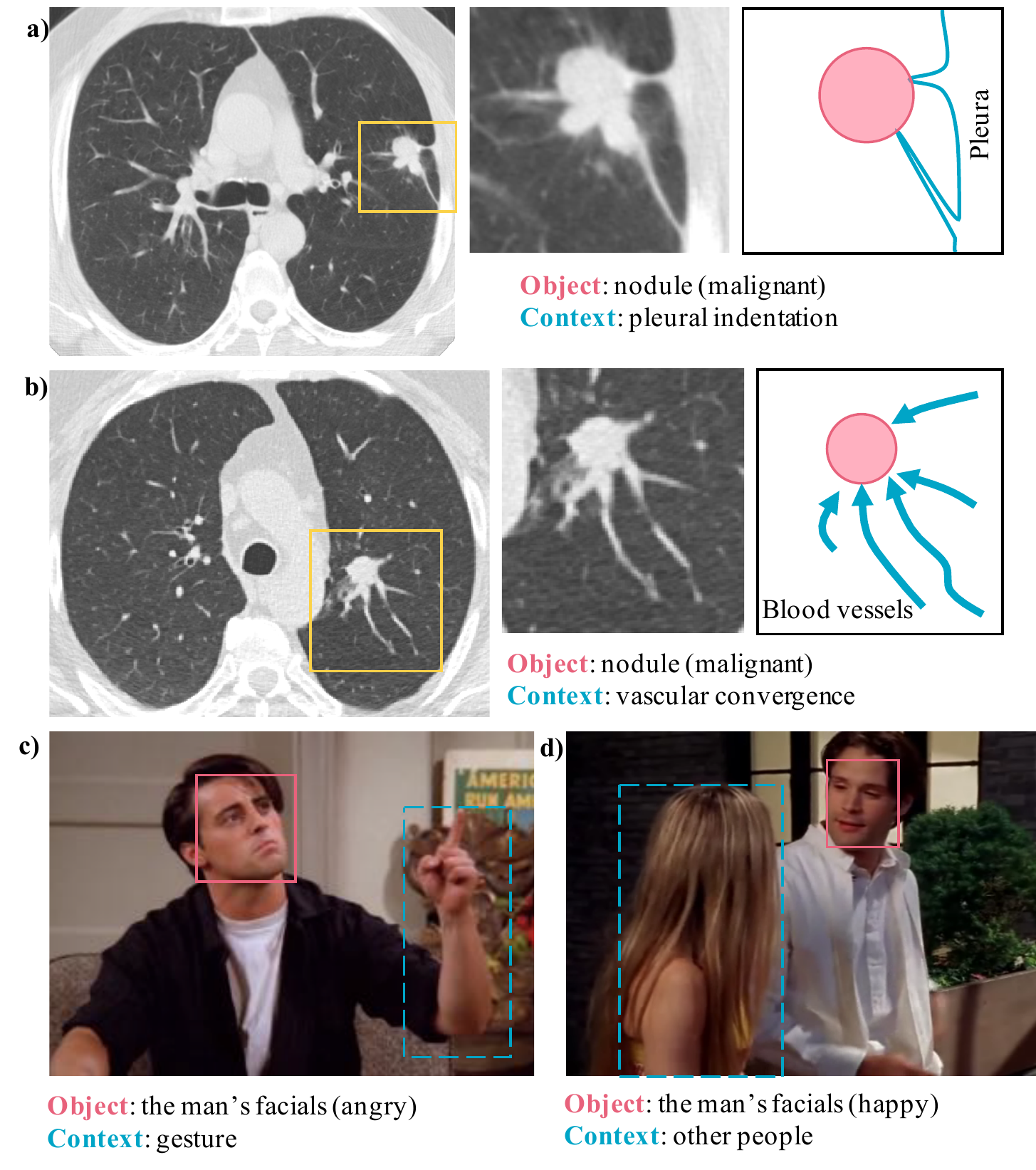}
\caption{Examples of objects and their correlated contextual features. (a) a malignant lung nodule that causes pleural indentation; (b) a malignant lung nodule with vascular convergence; (c) a man with angry facials and gestures; (c) a man with happy facials for meeting other people. Background knowledge for context in lung cancer is available in Appendix \ref{Sect_appendix}.}
\label{Fig_object_context_correlation}
\end{figure}

\IEEEPARstart{I}{n} visual recognition, the object can vary along with its relating situational factors such as visual scene, which are collectively referred to as \emph{context}. Therefore, context can provide helpful information to infer the object's states or properties. Consider the benignity/malignancy classification of lung nodule, the malignant nodules can cause structure deformation (\textit{e.g.} pleural indentation (Fig.~\ref{Fig_object_context_correlation}-a) \cite{Pathology_PleuralIndentation}, vascular convergence (Fig.~\ref{Fig_object_context_correlation}-b) \cite{Pathology_VasConverge}) through invasion and releasing chemical factors to their neighbors. As contextual signs, such deformation provides important cues for cancer diagnosis. Another common example is the recognition of facial emotion, where context can arise from the emotion sender (\emph{e.g.}, gesture (Fig.~\ref{Fig_object_context_correlation}-c) \cite{Psychology_Emotion(Context-Types-BodyGesture)}), other people (Fig.~\ref{Fig_object_context_correlation}-d) \cite{Psychology_Emotion(Context-Types-OtherPeople)}, and visual scene/objects (Fig.~\ref{Fig_exp_cross_att_all}-v,w,s,y) \cite{Psychology_Emotion(Context-Types-VisualScene-1)}. 

Despite its importance, capturing context can be very challenging. This is because contextual factors can dramatically change across instances, in terms of pattern and location. Again, consider two types of structure deformation context in the malignant nodule, the pleural indentation and the vascular convergence, as examples. The pleural indentation context appears as a tangent line and locates at the margin of the nodule, while the vascular convergence context appears as several diffuse vessels pointing toward the nodule and locates in the nodule's surrounding area. They differ significantly from each other in terms of pattern and location. Besides, the number of contextual factors can also vary a lot, \textit{e.g.} some nodules present a single structure deformation, while others can cause multiple ones (Fig.~\ref{Fig_exp_cross_att_all}-a,b) simultaneously.

As a result, it is intractable to annotate the ``context" for supervised learning. Without annotations, existing methods exploit the image label as the only source of supervision to learn the context, resulting in a large literature of different contextual tricks, including but not limited to features pyramid \cite{CVContext_SPP-Net-14,CVContext_Multi-Crop-17}, atrous convolution \cite{CVContext_DeepLab-v3plus-18}, and the attention modules \cite{CVContext_Non-Local-18,CVContext_A2-Net-18,CVContext_SE-Net-18,CVContext_Stand-Alone-19,CVContext_LR-Net-19,CVContext_AA-Conv-19,CVContext_GloRe-19,CVContext_ECA-Net-20,CVContext_LCT-20,CVContext_GCT-21}.
Although some of these methods showed performance improvements, they lacked careful modeling of context's properties, especially its relation to the object. This inherent weakness limits their potentials to locate accurate context and achieve better performance.

Recently, psychology researches \cite{Psychology_Congruent(Psy-Sci),Psychology_Congruent(face-body)} show that such an object-context relation is exploited by humans as common sense for recognition, as it can confuse human perceivers when the object and context are exposed incongruently. Inspired by this, we propose a novel \textbf{Context}ual \textbf{L}atent \textbf{G}enerative
\textbf{M}odel (\textbf{Context-LGM}), which models the object-context relation in a hierarchical perspective. Specifically, we firstly introduce a latent generative model coupled with a pair of correlated latent components to model the object and context, and embed their correlation via the generative processes. Then, guided by such modeling, we reformulate the \textbf{E}vidence \textbf{L}ower \textbf{BO}und (ELBO) of \textbf{V}ariational \textbf{A}uto-\textbf{E}ncoders (VAE) as objective function to infer object's and contextual features. Particularly, the posterior of contextual features is conditioned on the object, in order to take the object's information as a reference. Finally, to implement it, we introduce a contextual posterior Transformer \cite{Trans_Orig_Transformer}. With careful modifications, we show our Transformer can screen out environmental factors correlated to the object, and well handle the diversity of context.

To verify the utility of our Context-LGM, we evaluate it on two context-aware object recognition tasks, namely lung cancer prediction and emotion recognition. It yields that our method achieves state-of-the-art performance with significant margins. Besides, the visualization results qualitatively show that our method can locate interpretable context.  

In summary, our contributions are:
\begin{itemize}
    \item \textbf{Ideologically,} we are the \emph{first} to incorporate object-context relation into context-aware object recognition.
    \item \textbf{Methodologically,} we propose a novel latent generative model that embeds object-context relation in the generating process of two coupled latent variables.
    \item \textbf{Algorithmically,} we reformulate the ELBO of VAE as objective function and implement a contextual posterior Transformer to infer contextual features.
    \item \textbf{Experimentally,} we achieve state-of-the-art performance in two context-aware object recognition tasks.
\end{itemize}

The preliminary version of this work has been accepted by MICCAI-2021 \cite{CA_Net}. This paper extends the initial version in significant ways. \textbf{Firstly}, we propose a latent generative framework to carefully model context, especially its relation to the object. \textbf{Secondly}, we reformulate the ELBO of VAE and infer contextual features as a posterior distribution conditioned on the object. \textbf{Thirdly}, we modify the Transformer to implement this contextual posterior. \textbf{Finally}, we extend experiments to emotion recognition and provide more comprehensive results. 

The rest of the paper is organized as follows: Section \ref{Sect_relateworks} gives a brief review of related works; Section \ref{Sect_methodology} gives the detailed generative modeling, ELBO reformulating, and Transformer architecture of Context-LGM; Section \ref{Sect_experiment} provides quantitative and qualitative experimental results; Section \ref{Sect_conclusion} concludes the paper and discusses future works.

\section{Related works}
\label{Sect_relateworks}

\subsection{Context-Aware Object Recognition}

Many context-aware methods have been proposed in visual recognition, such as \cite{CVContext_SPP-Net-14,CVContext_Multi-Crop-17,CVContext_DeepLab-v3plus-18,CVContext_Non-Local-18,CVContext_A2-Net-18,CVContext_SE-Net-18,CVContext_Stand-Alone-19,CVContext_LR-Net-19,CVContext_AA-Conv-19,CVContext_GloRe-19,CVContext_ECA-Net-20,CVContext_LCT-20,CVContext_GCT-21}. The main focus lies in identifying context from complicated surrounding scenarios. To achieve this goal, existing methods follow the paradigm of using only image labels to capture context. Their typical designing tricks include features pyramid, atrous convolution, and attention modules.

The features pyramid \cite{CVContext_SPP-Net-14,CVContext_Multi-Crop-17} and atrous convolution \cite{CVContext_DeepLab-v3plus-18} were designed to enlarge the receptive fields and avoid missing context with extreme sizes or locations. Recently, the attention mechanism has become a new dominance for context modeling \cite{CVContext_Non-Local-18,CVContext_A2-Net-18,CVContext_SE-Net-18,CVContext_Stand-Alone-19,CVContext_LR-Net-19,CVContext_AA-Conv-19,CVContext_GloRe-19,CVContext_ECA-Net-20,CVContext_LCT-20,CVContext_GCT-21}. In these methods, various types of context attention implementations, including spatial-wise attention \cite{CVContext_Non-Local-18,CVContext_Stand-Alone-19,CVContext_LR-Net-19,CVContext_AA-Conv-19}, channel-wise attention \cite{CVContext_SE-Net-18,CVContext_LCT-20,CVContext_ECA-Net-20,CVContext_GCT-21}, spatial-channel wise attention \cite{CVContext_A2-Net-18}, and graph-based attention \cite{CVContext_GloRe-19}, were exploited to select label-correlated context and suppress other irrelevant backgrounds.

However, without further careful modeling of context's properties, especially its relation to the object, their learned context can suffer from large inaccuracy. Using even more designing tricks can not overcome these inherent obstacles they face. \textbf{In contrast}, our Context-LGM uses a pair of correlated latent variables to model the object and context, as well as the correlation between them. During inferring contextual features, we additionally incorporate the object's information as a reference. Such careful modeling of context and its relation to the object enable our methods capture better contextual representations, thus benefit the recognition.

\subsection{Latent Generative Model}

The latent generative model starts with a Bayesian Network, which introduces latent variables that are characterized via their generating processes. The simplest example is $\bm{z} \to \bm{x}$, in which $\bm{z}$ denotes the latent variables that generate $\bm{x}$. The goal is learning $p(x)$, which is however intractable for maximum likelihood based method if $\bm{x}$ is high-dimensional (\emph{e.g.}, $\bm{x}$ denotes image). The \textbf{V}ariational \textbf{A}uto-\textbf{E}ncoder (VAE) proposed the \textbf{E}vidence \textbf{L}ower \textbf{BO}und (ELBO) as a surrogate and tractable objective, by introducing the variational distribution $q(z|x)$ that is easy for sampling. 

Not only as a generator, the VAE can also be used to infer the latent representations from $x$. Specifically, it adopts the (variational) Encoder and Decoder, which respectively infer $z$ from $x$ and generate $x$ from $z$. There is a large literature in VAE to learn such meaningful representations \cite{higgins2016beta,chen2018isolating,li2021causal,sun2020latent}. Specifically, the \cite{higgins2016beta,chen2018isolating,ding2020guided} proposed to learn disentangled representations. Particularly, the \cite{sun2020latent,mahajan2020diverse} considered the supervised learning tasks and split the latent variables into two parts that were modeled differently. Benefited from its tractability and the ability to infer latent variables, we adopt the VAE framework to model the correlation between object and context via their latent generating processes. We then reformulate the ELBO based on the corresponding Bayesian Network. To the best of our knowledge, we are the \emph{first} to model contextual features and exploit its relation to the object of interest during inference. Our experimental results will show that such modeling can significantly improve the performance and learn interpretable context.

\subsection{Transformer}

Transformer \cite{Trans_Orig_Transformer} was initially proposed in natural language processing (NLP). Equipped with its unique multi-head attention in the self/cross-attention block, it can well model both the short-term and long-term dependencies among word tokens and thus achieves remarkable performance on various NLP tasks. Recently, Transformer and its attention mechanism have been introduced into visual relationships modeling, such as pixel-level dependencies in image classification \cite{Trans_Arch_Vit-11,Trans_Arch_DeiT-12} and segmentation \cite{Trans_Seg_CM-SAtt-15,Trans_Seg_Pano-Seg-92,Trans_Seg_VisTR-153}, relationships among proposals in detection \cite{Trans_Det_DETR-13}, correlation among keypoints in pose estimation \cite{Trans_PoseMesh_PoseFormer,Trans_PoseMesh_ContextPose-Xiaoxuan}, and temporal correspondence in tracking and action recognition \cite{Trans_Meet_Track,Trans_ActRec_ST-T-164}.

In our scenario, we implement a Transformer to model object-context correlation and infer contextual representations. To fit our task, we make careful modifications to the original designs. Specifically, we inherit the multi-head attention in the cross-attention block but implement it in a spatial-wise masking manner. This modification allows us to use the learned object-context correlation as a spatial mask and screen out contextual factors that are correlated with the object. Besides, we remove the Encoder-Decoder structure that is specifically designed for sequence-to-sequence translation. We also reduce the number of stacked attention layers and remove all FFNs to improve computational efficiency.


\begin{figure*}[t]
\centering
\subfloat[Contextual Generative Modeling]{\includegraphics[width=3.5in]{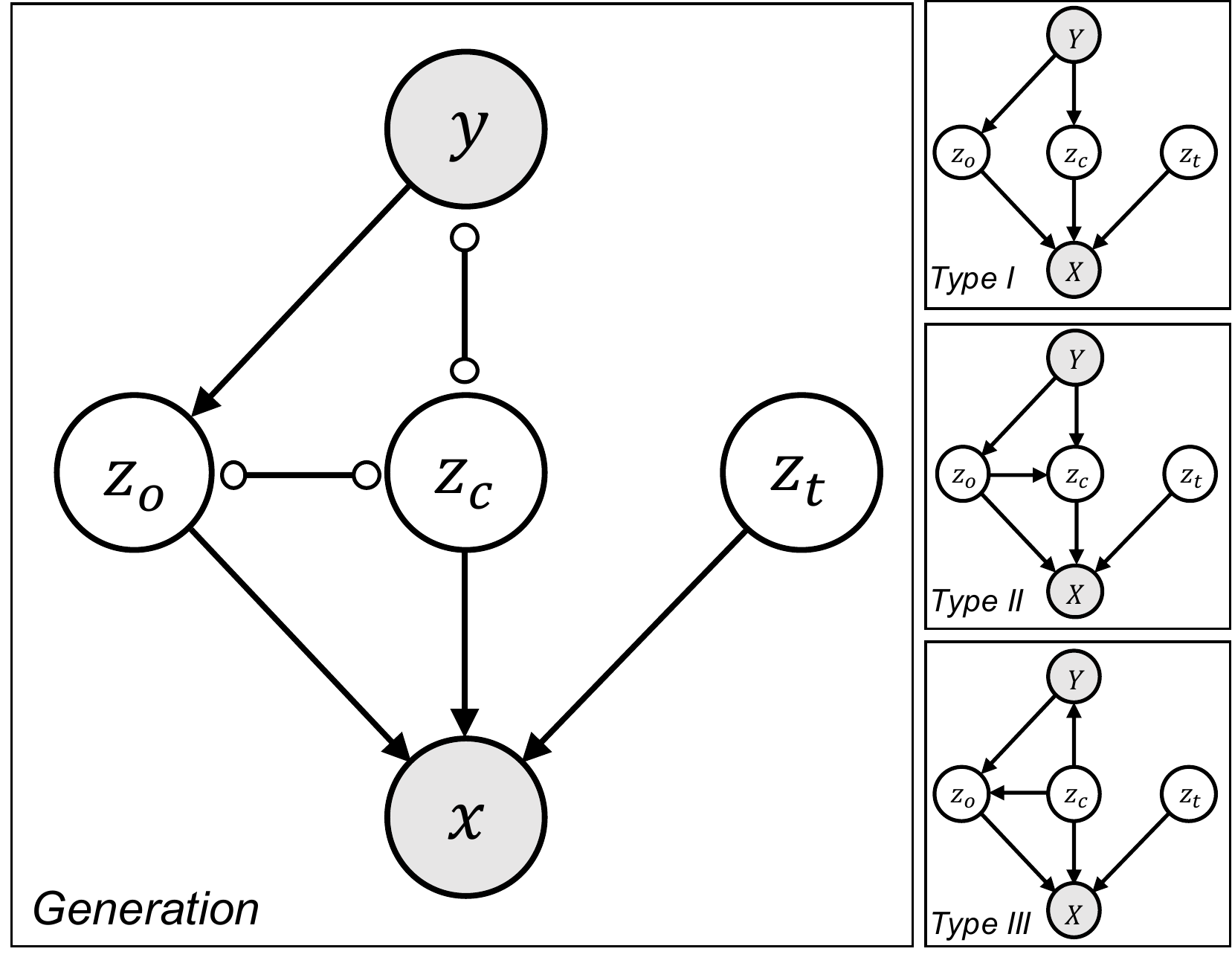}%
\label{fig1_first_case}}
\hfil
\subfloat[Latent Inference Process]{\includegraphics[width=2.625in]{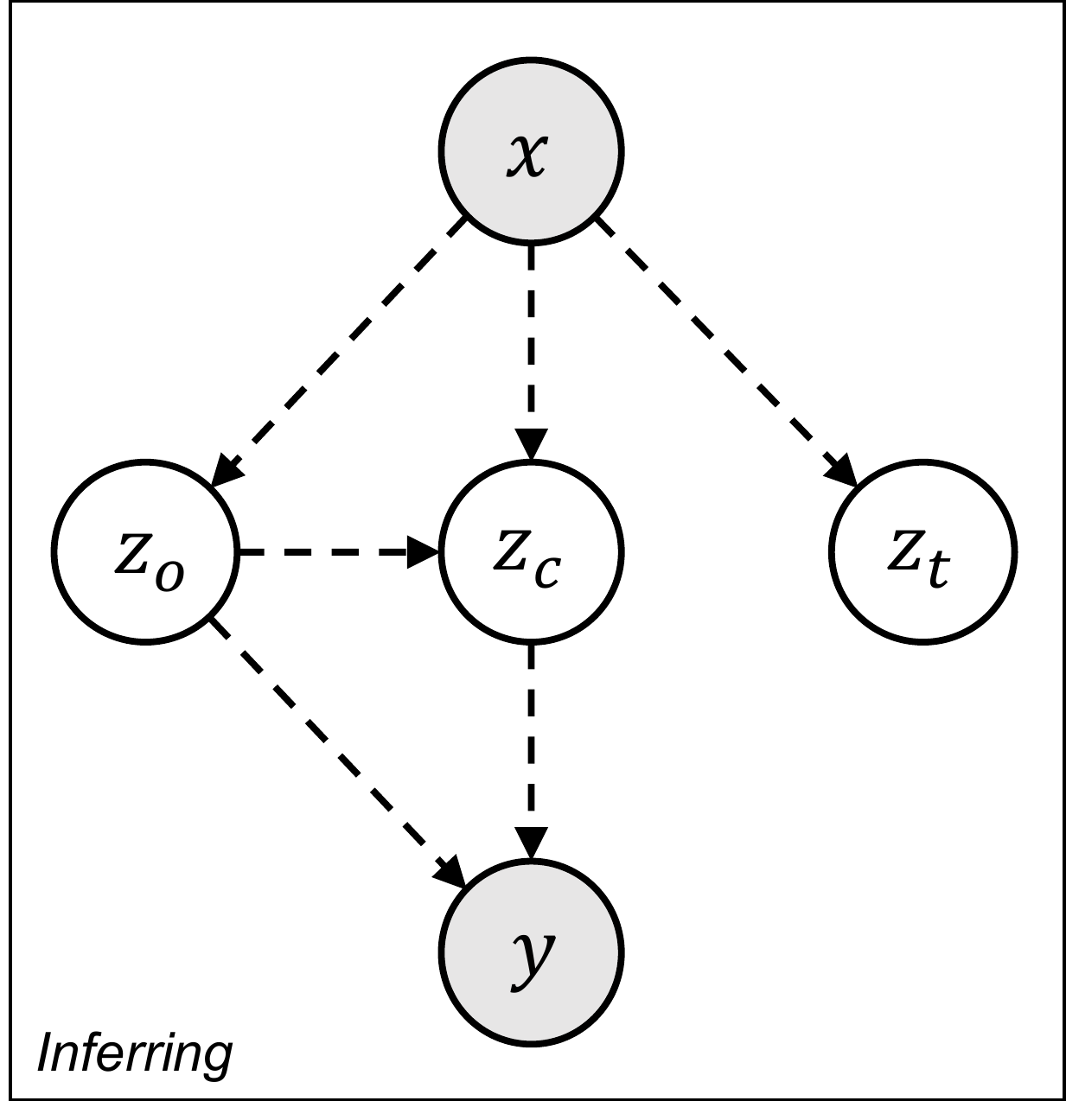}%
\label{fig2_second_case}}
\caption{(a) Contextual generative modeling. The \text{ o-o } mark denotes $\to$, $\gets$, or the missing link. We argue that an observed image is generated from three latent variables, \emph{i.e.}, $\bm{z_o}$, $\bm{z_c}$, and $\bm{z_b}$. These three variables respectively model the target object, context, and background features. More importantly, the object $\bm{z_o}$ and context $\bm{z_c}$ are correlated via their generative process. Specifically, they can be co-generated from $y$ (Type-I); Also, the object $z_o$ can also have direct influence on generating context $z_c$ (Type-II); Besides, the environmental context $z_c$ can generate the object's state $y$ and its features $z_o$ (Type-III). 
(b) Latent inference process. Given the observed image $\bm{x}$, we firstly infer object's latent $\bm{z_o}$ since the object is with annotations and easier to learn. Then, we infer contextual latent $\bm{z_c}$ from $\bm{x}$ using object' information $\bm{z_o}$ as a reference. Finally, $\bm{z_o}$ and $\bm{z_c}$ are jointly used to infer object's label $\bm{y}$. The white variable means unobserved variable, shaded variable means observed one.}


\label{Fig_latent_structure}
\end{figure*}

\section{Methodology}
\label{Sect_methodology}

\noindent \textbf{Problem Setup \& Notation.} Our goal is to predict object's state/label $y$ (\emph{e.g.}, benignity/malignancy or emotion) given an observed image $x$. The training data contains $\{x_i,y_i\}_i \overset{i.i.d}{\sim} p(x,y)$. In the graph of generative model, we use ``$o$" as a metasymbol to represent any kind of ends: '$>$', '$<$', and the empty mark,. That is, the $\bm{a} \text{ o-o } \bm{b}$ can represent $\bm{a} \to \bm{b}$, $\bm{a} \gets \bm{b}$, or the missing link between $\bm{a}$ and $\bm{b}$. The $\bm{a} \to \bm{b}$ means $\bm{b}$ is generated after $\bm{a}$. We use $\bm{x},x,\bm{X}$ to respectively denote the random variable, its instance and the matrix.

In this section, we introduce \textbf{Context}ual \textbf{L}atent \textbf{G}enerative \textbf{M}odel (\textbf{Context-LGM}) from a hierarchical perspectives of generative modeling (in section~\ref{Sect_methodology_1}), objective reformulation (in section~\ref{Sect_methodology_2}) and finally, the Transformer (in section~\ref{Sect_methodology_3}) for implementation. Specifically, in section \ref{Sect_methodology_1}, we firstly introduce the Bayesian Network and its encoded generative processes related to the object and context. Then, in section \ref{Sect_methodology_2}, we derive a reformulated ELBO as our objective function in the VAE framework to infer the object's and contextual features for prediction. Particularly, the posterior of contextual features is conditioned on the object's features, in order to take the object's information as a reference. Finally, in section \ref{Sect_methodology_3}, we implement a contextual posterior Transformer with a carefully modified cross-attention block to optimize the above posterior model.

\subsection{Contextual Generative Modeling}
\label{Sect_methodology_1}

We introduce three latent variables $\bm{z_o},\bm{z_c},\bm{z_b}$ to respectively model the object's, contextual and other background features. Together with the observed image $\bm{x}$ and object's state/label $\bm{y}$, their generating processes are illustrated by the Bayesian Network in Fig.~\ref{Fig_latent_structure}(a). As shown, the $\bm{z_o},\bm{z_c},\bm{z_b}$ all participate in generating the image $\bm{x}$, with each one responsible for corresponding source of variation in the whole image. Only $\bm{z_o},\bm{z_c}$ are related to the label $\bm{y}$. As an intuitive example, consider the man with angry facials and gesture in Fig.~\ref{Fig_object_context_correlation}(c). The $\bm{y}$ denotes his angry emotion, the $\bm{z_o}$ models his facial expressions, the $\bm{z_c}$ models his gesture context, and the $\bm{z_b}$ models the other emotion-irrelevant backgrounds such as the door, sofa, \textit{etc}.

More importantly, the object's features $\bm{z_o}$ is correlated to the contextual features $\bm{z_c}$ via their generating processes, as shown by the unblocked path between $\bm{z_o}$ and $\bm{z_c}$ in Fig.~\ref{Fig_latent_structure}(a). This path is composed of $\bm{y} \to \bm{z_o}$ (\emph{i.e.}, the object's state $\bm{y}$ generates its features $\bm{z_o}$) ,$\bm{z_o} \text{ o-o } \bm{z_c}$ and $\bm{z_c} \text{ o-o } \bm{y}$. It includes but not limited to the following possible types: 
\begin{itemize}
    \item $\bm{y} \to \bm{z_o}, \bm{y} \to \bm{z_c}$, missing link between $\bm{z_o}$ and $\bm{z_c}$ that corresponds to `Type I" in Fig.~\ref{Fig_latent_structure}(a). In this case, the object's state $\bm{y}$ co-generates object's features $\bm{z_o}$ and contextual features $\bm{z_c}$. For example, one can express his/her angry via simultaneous changes of facial expressions and body gestures. 
    \item $\bm{y} \to \bm{z_o}$, $\bm{y} \to \bm{z_c}$ and $\bm{z_o} \to \bm{z_c}$ that corresponds to ``Type II" in Fig.~\ref{Fig_latent_structure}(a). In this case, the generation of context $\bm{z_c}$ is not only influenced by object's state $\bm{y}$, but also directly by the object $\bm{z_o}$. For example, the structure deformation context is caused not only by the malignant nature of nodule, but also by its invasion pattern.
    \item $\bm{z_c} \to \bm{y}$, $\bm{z_c} \to \bm{z_o}$, and $\bm{y} \to \bm{z_o}$ that corresponds to ``Type III" in Fig.~\ref{Fig_latent_structure}(a). In this case, environmental context $\bm{z_c}$ influence the generation of object's state $\bm{y}$ and features $\bm{z_o}$. For example, scene objects and other people can influence the experience and expression of emotion \cite{Psychology_Emotion(Experi-Expres)}
\end{itemize}

Regardless of the types of path between $\bm{z_o}$ and $\bm{z_c}$, the label $\bm{y}$ and latent variables $\bm{z_o},\bm{z_c},$ and $\bm{z_b}$ obey the following properties: \textbf{i)} only $\bm{z_o},\bm{z_c}$ are related to $\bm{y}$; \textbf{ii)} $\bm{z_o}$ is related to $\bm{z_c}$. Mathematically speaking: 

\begin{equation}
\label{eq:1}
\begin{split}
    & \bm{y} \nperp \bm{z_o}|\bm{z_c}; \ \bm{y} \nperp \bm{z_c}|\bm{z_o}; \ \bm{z_o} \nperp \bm{z_c}; \  \\
    & \bm{z_b} \perp \bm{y}|\bm{z_o},\bm{z_c}; \ \bm{z_b} \perp \bm{z_o}; \ \bm{z_b} \perp \bm{z_c}; \
\end{split}
\end{equation}

These properties composite the cornerstone for the learning of object's and contextual representations, which we will introduce in the subsequent section.

\begin{figure*}[t]
\centering
\includegraphics[width=6.0in]{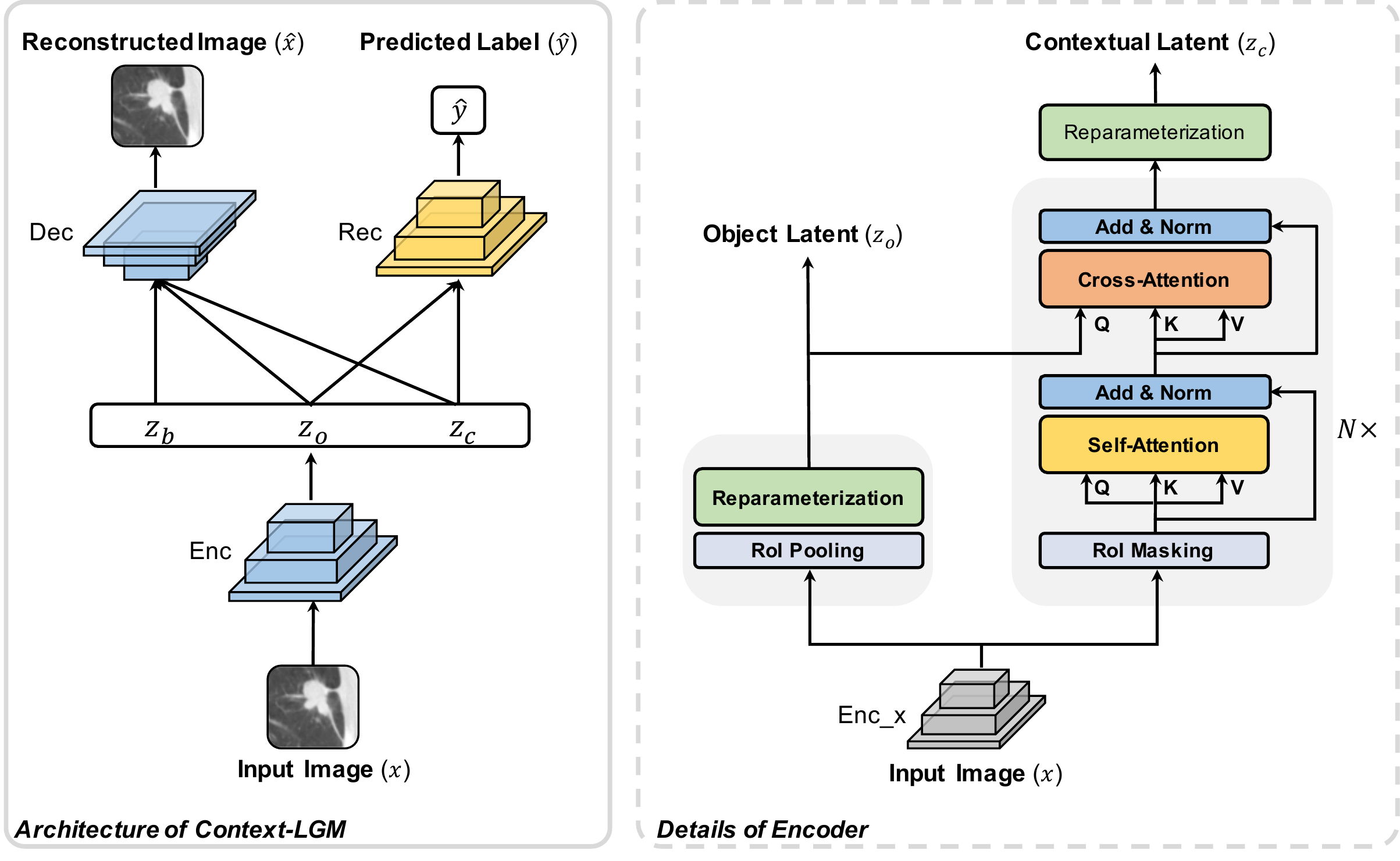}

\caption{Left: network structure of the proposed Context-LGM. Given the input image $x$, the Encoder (Enc) firstly extracts latent representations for the object ($z_o$) and its context ($z_c$). Then, the Decoder (Dec) takes $z_o$,$z_c$, and $z_b$ as input and outputs a reconstructed image $\hat x$;The Recognition Net (Rec) takes $z_o$ and $z_c$ as input and outputs the predicted label $\hat y$. Right: the structure of posterior model $q_\phi^c(z_o,z_c|x)$. It feeds $x$ into a backbone feature extractor (Enc\_x) and obtains deep feature maps $F$. To obtain $z_o$, $F$ is fed into a Region-Of-Interest (RoI) pooling layer followed by a reparameterization. To obtain $z_c$, the masked-RoI feature maps $F_{mask}$ is fed into a contextual posterior Transformer. The Transformer consists of N compositions of attention layers. Each attention layer contains a self-attention block and a cross-attention block. The self-attention block aims to mutually enhance features representation, while the cross-attention block takes the object's information $z_o$ to select correlated context $z_c$.}
\label{Fig_network_structure}
\end{figure*}
\subsection{Learning Method}
\label{Sect_methodology_2}

Guided by the Bayesian Network in Fig.~\ref{Fig_latent_structure}(a) and the properties derived in Eq.~\eqref{eq:1}, we introduce our method to infer $\bm{z_o}$ and $\bm{z_c}$ for prediction, by reformulating the ELBO of VAE. Specifically, the ELBO with $q_\phi(z|x)$ \, ($z:=[z_o,z_c,z_b]$ for simplicity) as variational distribution is:
\begin{equation}
\label{eq:elbo}
    \log p(y,x) \geq \log p_\psi(y|x) + \mathbb{E}_{q_\phi(z|x)} \log \frac{p_\psi(x|z)p(z)}{q_\phi(z|x)},
\end{equation}
where $q_\phi(z|x)$ is the variational distribution. 

We further approximate the $p_\psi(y|x)$ as:
\begin{align}
\label{eq:pred}
    p_\psi(y|x) & \approx \int p_\psi(y|z)q_\phi(z|x) dz. 
\end{align}

Then, the optimization for Eq.~\eqref{eq:elbo} involves the posterior model $q_\phi(z|x)$, the recognition model $p_\psi(y|z)$, the generation model $p_\psi(x|z)$, and the prior $p(z)$. The overall inference process and the network structure are respectively shown in Fig.~\ref{Fig_latent_structure} (b) and Fig.~\ref{Fig_network_structure}.

\noindent \textbf{Posterior Model.} For the posterior $q_\phi(z|x)$, we adopt the mean field approximation for the following factorization: 

\begin{equation}
    q_\phi(z|x)=q_\phi(z_o,z_c|x)q_\phi(z_b|x).
\label{Equation_posterior1}
\end{equation}
To leverage the object-context correlation, we incorporate the object's information as a reference during inferring $\bm{z_c}$: 
\begin{equation}
    q_\phi(z_o,z_c|x)=q_{\phi}^{c}(z_c|z_o,x)q_{\phi}^{o}(z_o|x),
\label{Equation_posterior2}
\end{equation}
with $q_{\phi}^{o}(z_o|x) \sim \mathcal{N}(\mu(x),\Sigma(x))$, and $q_{\phi}^{c}(z_c|z_o,x) \sim \mathcal{N}(\mu(z_o,x),\Sigma(z_o,x))$. To implement it, the object's posterior network $q_{\phi}^{o}(z_o|x)$ is parameterized by a backbone network followed by a Region-Of-Interest (RoI) pooling layer. The contextual posterior $q_{\phi}^{c}(z_c|z_o,x)$ is parameterized by the same backbone network, followed by a RoI masking layer and a contextual posterior Transformer. The detailed architecture of our Transformer will be introduced in the subsequent section.

\noindent \textbf{Recognition Model.} The recognition item has $p_\psi(y|z) = p_\psi(y|z_o,z_c)$, due to the independence between $y$ and $z_b$ in Eq. (\ref{eq:1}). The $p_\psi(y|z_o,z_c)$ is parameterized by a early concatenation of ($z_o,z_c$) followed by a three-layers fully connected classifier.

\noindent \textbf{Generation Model.} The generation model for image $p_\psi(x|z)$ is parameterized by a convolution layer followed by four-layer de-convolution layers.

\noindent \textbf{Objective Function.} With Eq.~\eqref{eq:pred},~\eqref{Equation_posterior1},~\eqref{Equation_posterior2}, the final loss is:
\begin{align}
    & \mathcal{L}_{\phi,\psi} = \frac{1}{n} \sum_{i=1}^{n} \left( -\lambda_1 \log{p_\psi(y_i|z_o,z_c)q_{\phi}^{c}(z_c|z_o,x_i)q_{\phi}^{o}(z_o|x_i)} \right. \nonumber \\
    & \left. + \lambda_2 D_{\mathrm{KL}}(q_\phi(z|x_i),p(z)) -  \lambda_3 \mathbb{E}_{q_\phi(z|x_i)}\log{p_\psi(x_i|z)} \right).
    \label{eq:loss}
\end{align}
where $n$ denotes samples number in our dataset. $\lambda_1$, $\lambda_2$, and $\lambda_3$ respectively denote ratios for recognition loss, KL divergence loss, and reconstruction loss.

\noindent \textbf{Training \& Test.} We optimize over $q_\phi(z|x)$, $p_\psi(x|z)$ and $p_\psi(y|z_o,z_c)$ by minimizing Eq.~\eqref{eq:loss} in the training stage. During test stage, we firstly infer $z_o$ via $q_\phi^{o}(z_o|x)$. Then, we infer $z_c$ via $q_{\phi}^{c}(z_c|z_o,x)$. Finally, we feed the inferred $z_o$ and $z_c$ into the recognition model $p_\psi(y|z_o,z_c)$ to predict $y$.

\subsection{Contextual Posterior Transformer}
\label{Sect_methodology_3}
\def\Q{\textbf{\textrm{Q}}}
\def\K{\textbf{\textrm{K}}}
\def\V{\textbf{\textrm{V}}}
\def\A{\textbf{\textrm{A}}}
\def\H{\textbf{\textrm{H}}}
\def\W{\textbf{\textrm{W}}}
\def\F{\textbf{\textrm{F}}}
\def\M{\textbf{\textrm{M}}}

\noindent \textbf{Overview.} Fig. \ref{Fig_network_structure} shows an overview of our Transformer. As we can see, we inherit the multi-head attention mechanism as our core implementation of object-context correlation, due to its ability on modeling various aspects of relationships. To adapt the Transformer well into our visual recognition task, we make the following modifications to the original designs: 

\begin{itemize}
    \item For the implementation of attention function in the cross-attention block, we replace the dot-product transformation attention (that is originally designed for source-target language translation) with a Hadamard product spatial-wise attention, to screen out those features that are highly correlated to the object.
    \item We remove the Encoder-Decoder structure that is specifically designed for the sequence-to-sequence translation but is not required in our task.
    \item we reduce the number of stacked attention layers and remove all FFNs. These can improve computational efficiency without loss of accuracy.
\end{itemize}

\noindent \textbf{Multi-Head Attention Function.} Attention function is the cornerstone in Transformer. Given the input query ${\Q}\subseteq \mathbb{R}^{N_q \times C}$, key $\K\subseteq \mathbb{R}^{N_k \times C}$, and value $\V\subseteq \mathbb{R}^{N_v \times C}$ ($N_q$,$N_k$,$N_v$ respectively denotes token length for query, key, and value, $C$ denotes channel dimension), the attention function is defined as a dot-product between affinity matrix $\A_{qk}=\Q \K^{T}$ and $\V$:

\begin{equation}
    \mathrm{Attention}(\Q,\K,\V) = \mathrm{Softmax}_{col}(\frac{\A_{qk}}{\tau})\V
\label{Equation_attention_function}
\end{equation}
where $\tau$ is a temperature parameter controlling the softmax distribution, $col$ denotes softmax is performed column wise. 

In our scenario, the contextual features can present multiple patterns/locations/quantities. So, we also extend the attention function in Eq.(\ref{Equation_attention_function}) into multi-head attention \cite{Trans_Orig_Transformer} to enhance its ability on modeling various aspects of contextual correlations. Specifically, the multi-head function is defined as: 

\begin{equation}
\begin{split}
    \mathrm{MultiHead}&(\Q,\K,\V) = \mathrm{Concat}(\H_1,\H_2,...,\H_{n_h}) \\
    &\H_i = \mathrm{Attention}(\Q\W_{i}^{Q},\K\W_{i}^{K},\V\W_{i}^{V})
\end{split}
\end{equation}
where $n_h$ denotes head numbers; $\W_{i}^{Q}\subseteq \mathbb{R}^{C_q \times C_{{q}'}}$, $\W_{i}^{K}\subseteq \mathbb{R}^{C_k \times C_{{k}'}}$, and $\W_{i}^{V}\subseteq \mathbb{R}^{C_v \times C_{{v}'}}$ are learned parameter metrics for query, key, and value;  $C_{{q}'}=\frac{C_q}{n_h}$,$C_{{k}'}=\frac{C_k}{n_h}$,$C_{{v}'}=\frac{C_v}{n_h}$. We share weight between $\W_{i}^{Q}$ and $\W_{i}^{K}$, as it has been suggested \cite{Trans_Meet_Track} to embed the (query,key) into the same space and help accurate correlation computing. 

\noindent \textbf{Self-Attention Block.} The self-attention block aims to mutually enhance feature maps representations. It takes the masked RoI feature maps $\F_{mask} \subseteq \mathbb{R}^{H \times W \times C}$ ($H$,$W$ denote spatial dimensions, and $C$ denotes channel dimension) as input. We reshape $\F_{mask}$ into $\mathbb{R}^{C \times N}$ ($N=H \times W$) by flattening the spatial dimensions and feed it in into the multi-head attention function. The output of the attention is added to the input feature maps as a residual item:
\begin{equation*}
    \F_{self} = \mathrm{Norm.}(\mathrm{MultiHead}(\F_{mask},\F_{mask},\F_{mask}) + \F_{mask})
\end{equation*}
where $\mathrm{Norm.}$ denotes a normalization layer and $\F_{self}$ denotes output of self-attention block.

\noindent \textbf{Cross-Attention Block.} As our key implementation of the object-context correlation, the cross-attention block leverages the object's information as guidance in learning the contextual representations. 

Specifically, it takes both object's latent $z_o$ and the output of self-attention block $\F_{self}$ as input. In its multi-head attention function, the object's latent $z_o$ is used as query and $\F_{self}$ is used as key and value, such that feature tokens in $\F_{self}$ with high correlation to $z_o$ can be enhanced and those with low correlations can be suppressed:
\begin{equation*}
    \F_{cross} = \mathrm{Norm.}(\mathrm{MultiHead}(z_o,\F_{mask},\F_{mask}) + \F_{mask})
\end{equation*}

The contextual latent $z_c$ is then estimated by a reparameterization on $\mathbf{F}_{cross}$. 

Note that originally, the attention in Eq.~\eqref{Equation_attention_function} is implemented by a dot-product between affinity \A \ and value \V , such that the source sentence (\V) can be transformed into target language (\Q) domain according to words correspondence (\A). However, as mentioned earlier, our cross-attention block aims to emphasize environmental factors with high correlation to the object and suppress those with low correlation. Hence, we implement our cross-attention in a spatial-wise attention manner:
\begin{equation}
    \mathrm{Attention}(\Q,\K,\V)=\mathrm{Softmax}_{col}(\frac{\M_{qk}}{\tau})\odot \V
\end{equation}
with $\M_{qk}:=\sum_{row}A_{qk}$, as the spatial-wise object-context correlation matrix.
``$\sum_{row}$" means the summation is performed row-wise and $\odot$ denotes a Hardmard product.

During the Hardmard product, the left spatial correlation item is a $N_k\times 1$ matrix, and the right value item \V \ is a $N_k\times C$ matrix. The spatial correlation is shared among different channel dimensions.

\section{experiments}
\label{Sect_experiment}
In this section, we report quantitative and qualitative results of the proposed Context-LGM on two context-aware object recognition tasks, \textit{i.e.} lung cancer prediction, and emotion recognition. Correspondingly, in lung cancer prediction, the contextual features include structure deformations/attachments by nodules; In emotion recognition, they include gestures, scene/objects, or other people.

\begin{figure}[t]
\centering
\includegraphics[width=3.5in]{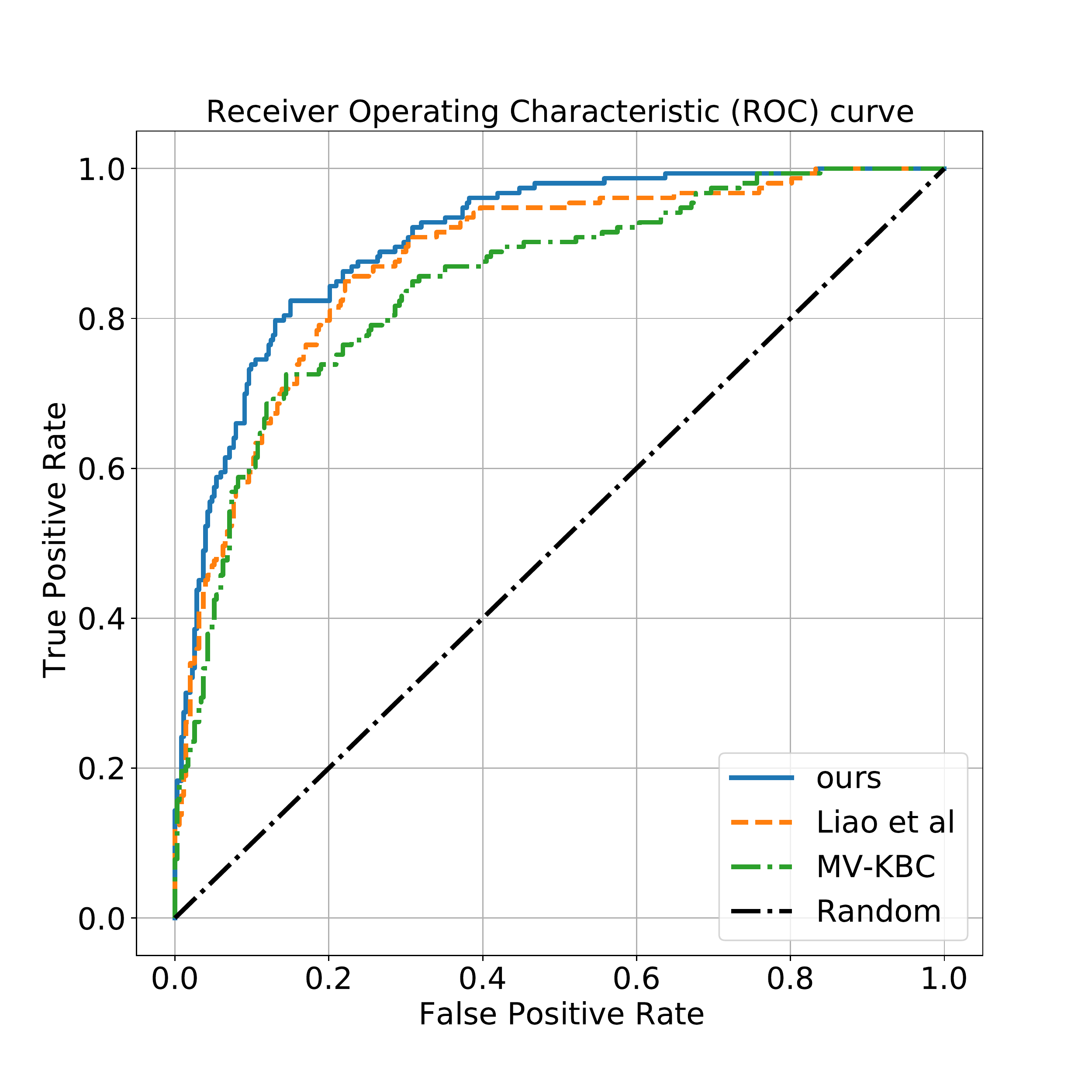}
\caption{ROC curves of different methods in lung cancer prediction task. Note that detailed prediction probability for each data point is required to draw ROC curves. So only baseline methods with official training codes or model weights are included for fairness.}
\label{Fig_dsb2017_roc}
\end{figure}
\begin{table*}[t]
\renewcommand{\arraystretch}{1.3}

\centering
\begin{tabular}{c c c c c}
\hline
Method & Publish & Context &AUC($\%$) $\uparrow$ & LogLoss $\downarrow$ \\
\hline
grt123 & \ & $\checkmark$ & - & .3998 \\
J. de Wit \& D. Hammack & \ & $\checkmark$ & - & .4012 \\
Aidence & Competition & - & - & .4013 \\
qfpxfd & \ & - & - & .4018 \\
Pierre Fillard & \ & - & - & .4041 \\
\hline
MV-KBC \cite{Xie-TMI-2018} & TMI 18 & \XSolidBrush & 84.67 & .4887 \\
Ozdemir \textit{et al.} \cite{Ozdemir-TMI-2019} & TMI 19 & $\checkmark$ & 86.90 & - \\
Liao \textit{et al.} \cite{Liao-TNNLS-2019} & TNNLS 19 & $\checkmark$ & 87.00 & .3989 \\

\textbf{ours} & \ & $\checkmark$ & \textbf{90.24} $\pm$ 0.16 & \textbf{.3836} $\pm$ .0065 \\ 
\hline
\end{tabular}
\caption{Comparison results over other methods in lung cancer prediction task. `Context' denotes whether context information is contained in the input image. `-' means unavailable. Our result is reported in mean $\pm$ std. Standard variance for baseline methods are not reported in their papers.}
\label{Tab_dsb2017_leaderboard}
\end{table*}
\begin{table*}[t]
\renewcommand{\arraystretch}{1.3}
\centering
\begin{tabular}{c c c c c}
\hline
Backbone & Method & Publish & Context & Acc($\%$) $\uparrow$ \\
\hline
\multirow{4}*{5-Layers CNN} & CAER-Net \cite{CAER} & ICCV 19 & $\checkmark$ &  77.83* \\ 
\ & Jaiswal \textit{et al.} \cite{CAER-Jaiswal} & UPCON 2020 & $\checkmark$ & 81.00 \\
\ & Zeng \textit{et al.} \cite{CAER-Zeng} & CSRSWTC 2020 & $\checkmark$ & 81.31 \\
\ & ours & \ & $\checkmark$ & 85.30 $\pm$ 0.07 \\
\hline

\multirow{5}*{ResNet-18} & SIB-Net \cite{CAER-Li-IJCB} & IJCB 21 & $\checkmark$ &  74.56 \\
\ & Li \textit{et al.} \cite{CAER-Li-TAC} & TAC 21 & $\checkmark$ &  84.82 \\
\ & EfficientFace \cite{CAER-Zhao} & AAAI 21 & \XSolidBrush &  85.87 \\
\ & MA-Net \cite{MA-Net} & TIP 21 & \XSolidBrush &  88.42 \\
\ & \textbf{ours} & \ & $\checkmark$ &\textbf{91.36} $\pm$ 0.05 \\
\hline
\end{tabular}
\caption{Comparison results in emotion recognition task. EfficientFace \cite{CAER-Zhao} used ResNet-50 as backbone. * denotes our re-training result, original result is 73.51\%. `context' denotes whether considering context information. Our result is reported in mean $\pm$ std. Standard variance for baseline methods are not reported in their papers.}
\label{Tab_caer2017_leaderboard}
\end{table*}
\begin{table}[!t]
\renewcommand{\arraystretch}{1.0}

\centering
\begin{tabular}{ c c | c c | c c }
\hline
\multicolumn{2}{c|}{vanilla} & \multicolumn{2}{c|}{ours} & DSB2017 & CAER-S \\

object & context & VAE & Transformer &  AUC($\%$) $\uparrow$ & Acc($\%$) $\uparrow$ \\
\hline
$\checkmark$ & \ & \ & \ & 88.06 $\pm$ 0.67 & 71.89 $\pm$ 0.26 \\
$\checkmark$ & $\checkmark$ & \ & \ & 88.72 $\pm$ 0.15 & 76.96 $\pm$ 0.17 \\
$\checkmark$ & $\checkmark$ & $\checkmark$ & \ & 88.88 $\pm$ 0.02 & 80.27 $\pm$ 0.14 \\
$\checkmark$ & $\checkmark$ & $\checkmark$ &  $\checkmark$ & \textbf{90.24} $\pm$ 0.16 & \textbf{85.30} $\pm$ 0.07 \\

\hline
\end{tabular}
\caption{Ablative study of our method. Vanilla object/context mean they are directly learned via cross entropy.}
\label{Tab_self_ablative}
\end{table}
\begin{table*}[!t]
\renewcommand{\arraystretch}{1.3}

\centering
\begin{tabular}{ c c c c c }
\hline
Method & Publish & Applications & DSB2017(AUC) & CAER-S(Acc) \\
\hline
SPP \cite{CVContext_SPP-Net-14} & ECCV 14 & image classification, object detection & 88.63 $\pm$ 0.27 & 77.10 $\pm$ 0.16 \\

PSP \cite{CVContext_PSP-17} & CVPR 17 & scene parsing & 88.76 $\pm$ 0.04 & 84.25 $\pm$ 0.11 \\
Multi-Crop* \cite{CVContext_Multi-Crop-17} & PR 17 & image classification & 89.02 $\pm$ 0.15 & 83.23 $\pm$ 0.41 \\

DeepLab v3+ \cite{CVContext_DeepLab-v3plus-18} & ECCV 18 & semantic segmentation & 88.94 $\pm$ 0.13 & 80.90 $\pm$ 0.22 \\
$A^{2}$-Net \cite{CVContext_A2-Net-18} & NeurIPS 18 & image/video classification & 89.28 $\pm$ 0.05 & 82.89 $\pm$ 0.35 \\
Context-Enc \cite{CVContext_Context-Enc-18}& CVPR 18 & semantic segmentation, image classification & 89.23 $\pm$ 0.09& 83.84 $\pm$ 0.02 \\
Non-Local* \cite{CVContext_Non-Local-18} & CVPR 18 & image/video classification, instance segmentation, pose estimation & 89.02 $\pm$ 0.24 & 82.36 $\pm$ 0.41 \\
SE-Net* \cite{CVContext_SE-Net-18} & CVPR 18 & image classification & 89.28 $\pm$ 0.31& 83.52 $\pm$ 0.16\\

Stand-Alone* \cite{CVContext_Stand-Alone-19} & NeurIPS 19 & image classification, object detection & 89.40 $\pm$ 0.27 & 83.45 $\pm$ 0.93 \\
LR-Net* \cite{CVContext_LR-Net-19} & ICCV 19 & image classification & 89.12 $\pm$ 0.25 & 84.46 $\pm$ 0.18 \\
Asy Non-Local \cite{CVContext_Asy-Non-Local-19} & ICCV 19 & semantic segmentation & 89.07 $\pm$ 0.32 & 79.78 $\pm$ 0.69 \\
AA-Conv* \cite{CVContext_AA-Conv-19} & ICCV 19 & image classification, object detection & 89.24 $\pm$ 0.32 & 82.58 $\pm$ 0.13 \\
GloRe* \cite{CVContext_GloRe-19} & CVPR 19 & image classification, semantic segmentation, action recognition & 88.83 $\pm$ 0.68 & \underline{84.65} $\pm$ 0.97 \\
DA-Net \cite{CVContext_DA-Net-19} & CVPR 19 & scene parsing & 88.94 $\pm$ 0.38 & 82.30 $\pm$ 0.44 \\

LCT* \cite{CVContext_LCT-20}& AAAI 20 & image classification, object detection & 89.06 $\pm$ 0.43 & 84.34 $\pm$ 0.24 \\
CaC-Net \cite{CVContext_CaC-Net-20}& ECCV 20 & semantic segmentation & 89.37 $\pm$ 0.02 & 77.86 $\pm$ 0.49 \\
Att Deeplab v3+ \cite{CVContext_Att-Deeplab-v3-plus-20}& ECCV 20 & lesion segmentation & 89.01 $\pm$ 0.11 & 82.23 $\pm$ 0.12 \\
ECA-Net* \cite{CVContext_ECA-Net-20}& CVPR 20 & image classification, object detection, instance segmentation & \underline{89.41} $\pm$ 0.14 & 83.99 $\pm$ 0.59 \\

CG-Net \cite{CVContext_CG-Net-21} & TIP 21 & semantic segmentation & 88.90 $\pm$ 0.71 & 83.16 $\pm$ 0.11 \\
RCAM \cite{CVContext_RCAM-21} & CVPR 21 & glass surface detection & 88.66 $\pm$ 0.10 & 77.91 $\pm$ 0.54 \\
CGT* \cite{CVContext_GCT-21} & CVPR 21 & image classification, object detection & 88.70 $\pm$ 0.34 & 83.91 $\pm$ 0.58\\

\hline
\textbf{ours} & \ & \ & \textbf{90.24} $\pm$ 0.16 & \textbf{85.30} $\pm$ 0.07 \\
\hline
\end{tabular}
\caption{Results of replacing our contextual posterior Transformer with other non object-context correlation context modeling methods when estimate contextual latent $z_c$. * behind a method's name denotes the method is originally used as a context block inside the convolutional backbone. results are reported in mean $\pm$ std.}
\label{Tab_cross_ablative}
\end{table*}
\begin{table}[!t]
\renewcommand{\arraystretch}{1.3}

\centering
\begin{tabular}{ c | c c }
\hline
layers number & DSB2017(AUC) & CAER-S(Acc) \\
\hline
1 & 89.95 $\pm$ 0.42 & \textbf{85.30} $\pm$ 0.07 \\
2 & \textbf{90.24} $\pm$ 0.16 & 84.13 $\pm$ 0.12 \\
3 & 89.20*$\pm$ 0.04 & 84.41 $\pm$ 0.12 \\
\hline
\end{tabular}
\caption{Influence of Transformer layers number. * denotes using NVIDIA APEX library to conduct FP16/FP32 mixed precision training due to GPU memory constraints.}
\label{Tab_hyper_paras_layernum}
\end{table}
\begin{table}[!t]
\renewcommand{\arraystretch}{1.3}

\centering
\begin{tabular}{ c | c c }
\hline
heads number & DSB2017(AUC) & CAER-S(Acc) \\
\hline
1 & 89.98 $\pm$ 0.05 & 85.08 $\pm$ 0.05 \\
2 & 90.00 $\pm$ 0.06 & \textbf{85.30} $\pm$ 0.07 \\
4 & \textbf{90.24} $\pm$ 0.16 & 84.97 $\pm$ 0.13 \\
8 & 89.26*$\pm$ 0.02  & 84.92 $\pm$ 0.12 \\
\hline
\end{tabular}
\caption{Influence of Transformer heads number. * denotes using NVIDIA APEX library to conduct FP16/FP32 mixed precision training due to GPU memory constraints.}
\label{Tab_hyper_paras_headnum}
\end{table}

\begin{figure*}[t]
\begin{minipage}[d]{0.1\linewidth}
\centering
\includegraphics[width=7.2in]{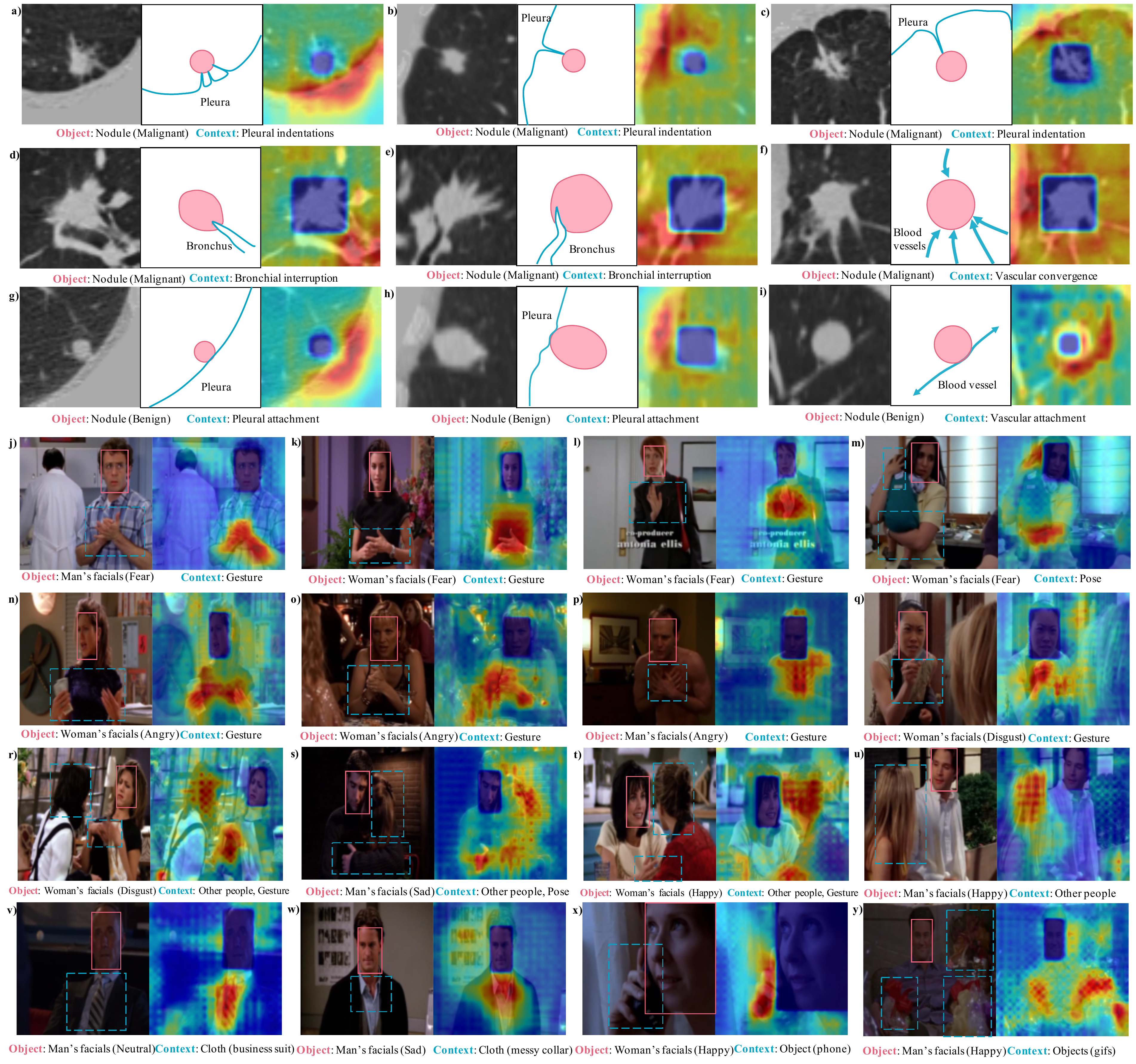}
\end{minipage}
\caption{Visualization of the learned object-context correlation matrices. For each example, the left image shows the input $x$ to our Context-LGM, and the right image shows the computed correlation matrix. Object regions in the correlation matrix are with zero response. This is because we mask them in $F_{mask}$ to avoid interference from object-object correlation.}
\label{Fig_exp_cross_att_all}
\end{figure*}

\begin{figure*}[ht]
\centering
\includegraphics[width=7.0in]{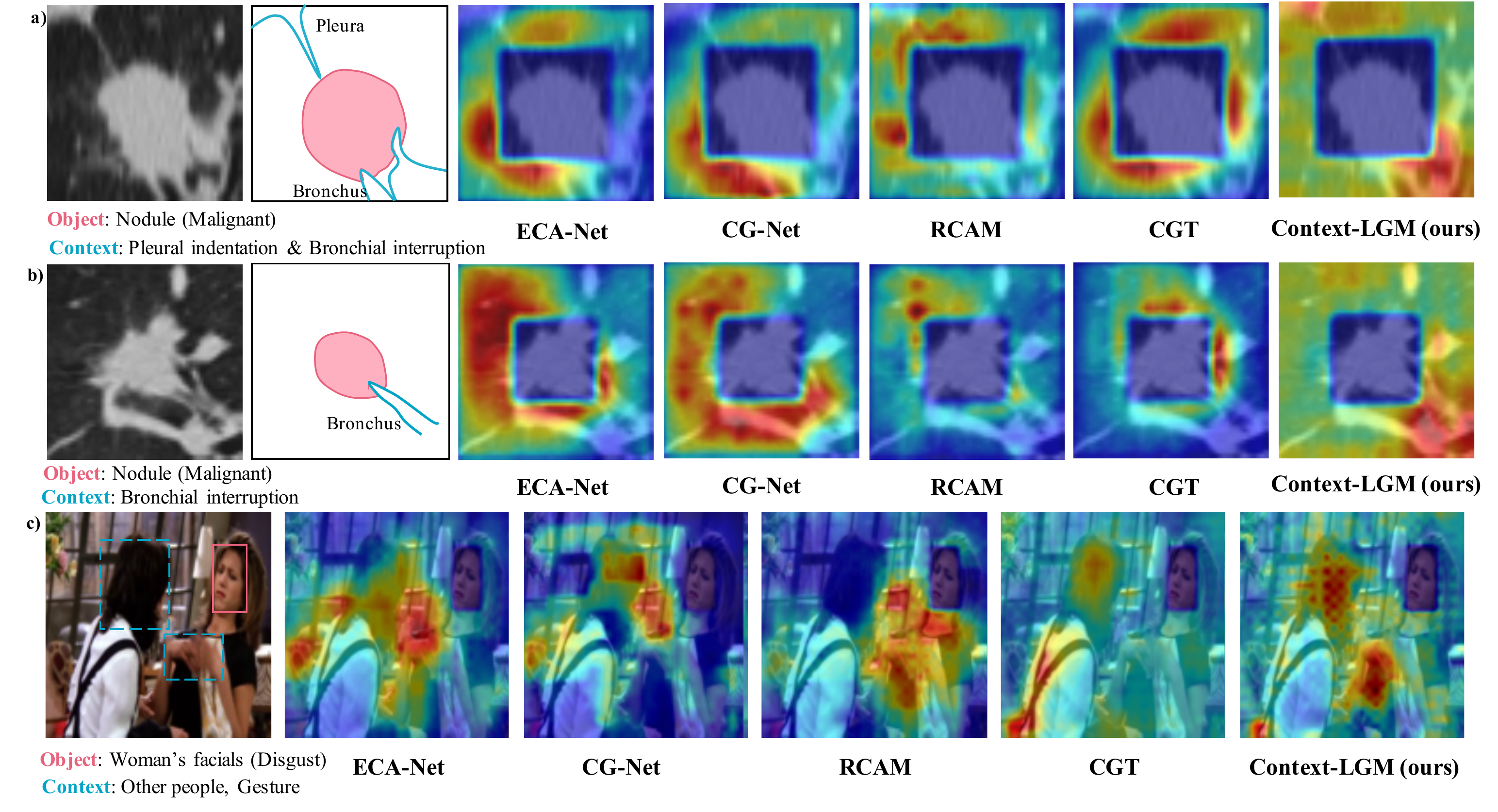}
\caption{Visualization of learned context in different methods.} 
\label{Fig_exp_visu_compare}
\end{figure*}

\begin{figure*}[ht]
\centering
\includegraphics[width=6.8in]{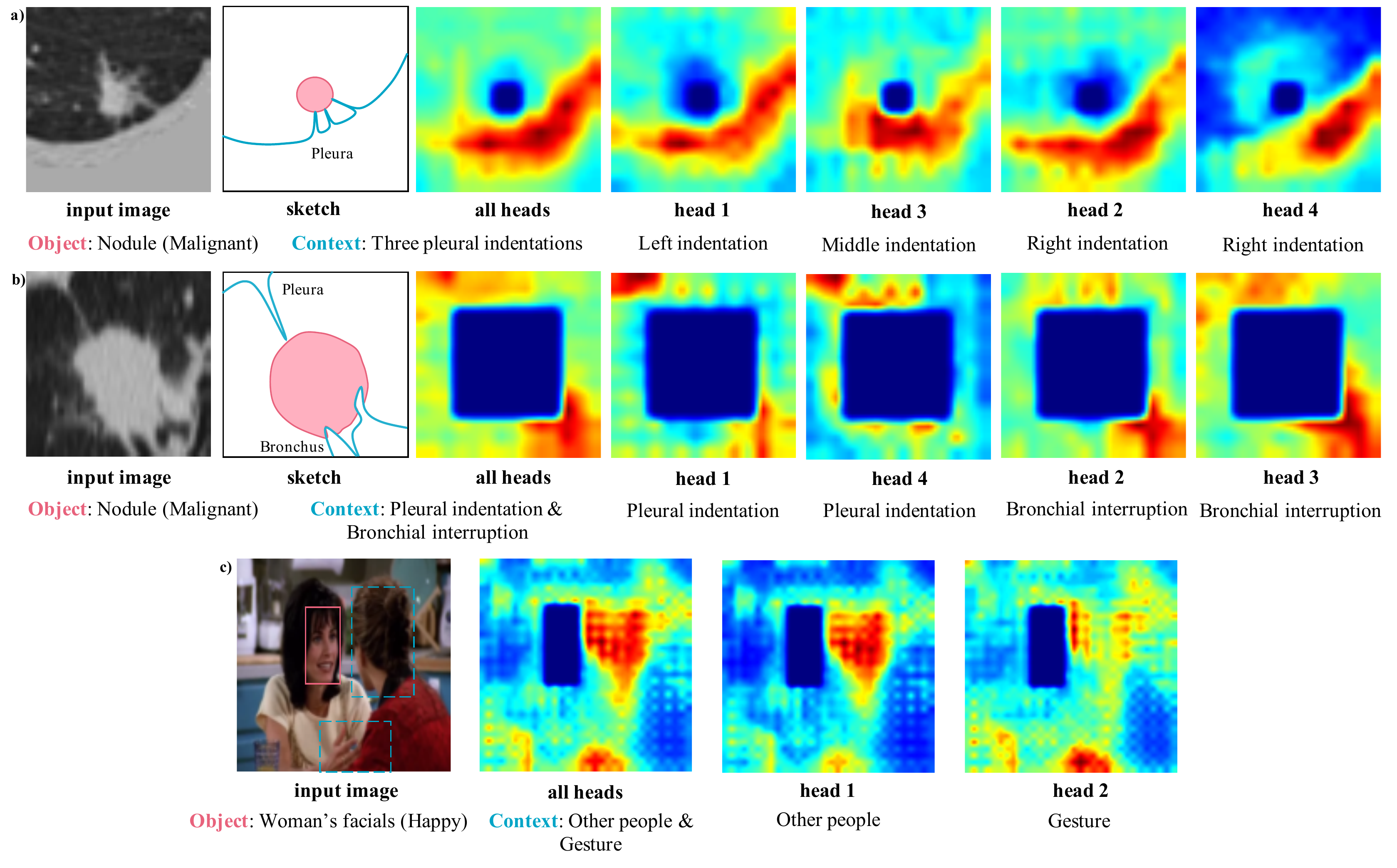}
\caption{Comparison of correlation matrices in different attention heads.}
\label{Fig_exp_cross_att_head}
\end{figure*}

\subsection{Datasets and Evaluation Metrics}

\noindent \textbf{Lung Cancer Prediction.} The Data Science Bowl (DSB) 2017 dataset \cite{KaggleDSB2017} in Kaggle's competition is used. This dataset provides a pathologically confirmed lung cancer label for each patient. There are 1397, 198, and 506 patients in the training, validation, and test set, respectively. We report official evaluation metrics including the Receiver Operating Characteristic (ROC) curve, the Area Under ROC curve (AUC), and the Log Loss (also known as the cross-entropy loss). The various types of contextual features (\emph{i.e.}, structure distortions, attachments) and its relation to the nodule is illustrated in Fig.~\ref{Fig_pathological_background1}(b-g).

\noindent \textbf{Emotion Recognition.} The Context-Aware Emotion Recognition (CAER-S) dataset \cite{CAER} is used. This dataset contains 48,971 images for training and 20,954 images for validation. Each image is assigned to one of the seven emotion categories (neutral, happy, sad, surprise, fear, disgust, and angry) based on people's facial expressions and context. We report its official evaluation metric, \textit{i.e.} Top-1 Accuracy (Acc). The context specifically denotes gestures, scene/objects, and other people, as marked by the dash blue boxes in Fig.~\ref{Fig_exp_cross_att_all} (i-y). 

\subsection{Implementation Details}
For both tasks, we implement SGD for optimization, with momentum set to $0.9$ and weight decay to $0.0001$. 

\noindent \textbf{Lung Cancer Prediction.} Given CT image $x$ and cancer label $y \subseteq \{0,1\}$, we implement an off-the-shell nodule detector \cite{Liao-TNNLS-2019} to detect all nodules $\{N_1,N_2,...,N_n\}$ in the patient. For each nodule $N_i$, an $96\times96\times96 \mathrm{mm}^{3}$ patch containing both nodule and surroundings is cropped and feed into Context-LGM to predict a malignancy score $\hat y_i$. Based on all nodules' malignancy probabilities $\{\hat y_1,\hat y_2, ..., \hat y_n\}$, the predicted cancer probability is defined by $\hat y = 1-\prod_{i=0}^{n} \hat y_i$.

Our backbone is a 3D U-Net. Data pre-processing includes pixel space re-sampling ($1\times1\times1 \mathrm{mm}^{3}$), intensity normalization (window center HU$=-600$, window width HU$=1600$), and lung segmentation. Data augmentations include random flipping, resizing, rotation and shifting. Alternative training on classification and detection is used to alleviate over-fitting. 

In the contextual posterior Transformer, short cuts between each attention layer are used to alleviate gradients vanishing as suggested by \cite{DenseNet}. Channel dimension $C$ is $128$. The heads number is set to $4$, the layers number is set to $2$, $\tau$ is set to $30$. The training takes $60$ epochs, with the learning rate set to $0.01$ and decreases by a factor of $0.1$ at epoch $20,35$. We multiply the learning rates for modules other than the backbone by a factor of $1.25$ in the first $20$ epochs to promote faster convergence. Due to GPU memory constraints, the batch size is set to $12$. The $\lambda_1,\lambda_2,\lambda_3$ in Eq.~\eqref{eq:loss} for recognition loss, KL divergence loss, and reconstruction loss are all set to $1.0$. It takes eight hours to train on two Nvidia Tesla V100 GPUs.

\noindent \textbf{Emotion Recognition.} We first crop the target face in each image and resize it into $96 \times 96$. We then mask the target face (by zero paddings) and resize its surrounding areas into $140\times160$. We tried various backbones: 5-layers CNN adopted in \cite{CAER,CAER-Jaiswal,CAER-Zeng} and ResNet-18 adopted in \cite{CAER-Li-IJCB,CAER-Li-TAC,MA-Net,CAER-Zhao}. Data augmentations include random cropping ($128\times128$) and flipping. Our contextual posterior Transformer takes fusion of multi-levels backbone feature maps as input. Channel dimension $C$ is $256$, heads number is set to $2$, layer number is set to $1$, and $\tau$ is set to $30$. The training takes $85$ epochs, with the learning rate set to $0.01$ and decrease by a factor of $0.1$ at epoch $55$. The batch size is set to $128$. The $\lambda_1,\lambda_2,\lambda_3$ in Eq.~\eqref{eq:loss} for recognition loss, KL divergence loss, and reconstruction loss are set to $1.0$, $0.1$, $1.0$, respectively. It takes five hours to train on one Nvidia Tesla V100 GPU.

\subsection{Results Analysis}

\noindent \textbf{Lung Cancer Prediction.} We compare our method with baselines from Kaggle's competition leader board and other state-of-the-art methods. For baselines from the leader board, team `grt123' and `J. de Wit \& D. Hammack' contained contextual information in their input image. Specifically, `grt123' extracted nodule's and contextual features together by a $2\times2$ pooling, `J. de Wit \& D. Hammack' used multiple tasks framework to jointly learn different attributes and nodule's malignancy. Whether using contextual information was not described in the competition reports of team `Aidence', `qfqxfd', and `Pieere Fillard'. `Aidence' used a multiple tasks learning framework. `qfqxfd', and `Pieere Fillard' adopted boosting methods to ensemble models trained with different settings. For other methods, MV-KBC \cite{Xie-TMI-2018} did not consider contextual information. They designed a 2D multi-view network to jointly learn the nodule's texture, internal characteristics, and margin. Ozdemir \textit{et al.} \cite{Ozdemir-TMI-2019} used a 3D probabilistic model, where nodule's and contextual features were extracted together by a global average pooling. Liao \textit{et al.} \cite{Liao-TNNLS-2019} was the extended version of team `grt123' with refined optimization strategy.

Numerical results are reported in Tab. \ref{Tab_dsb2017_leaderboard}. As we can see, our Context-LGM reaches an AUC of 90.24\%, outperforming the state-of-the-art baseline by 3.2\%. The distribution of predicted cancer probability is shown by the ROC curve in Fig. \ref{Fig_dsb2017_roc}, it can be further concluded that our method achieves better performance under almost all TP/FP rate settings. These results provide a strong verification of our method's effectiveness.

\noindent \textbf{Emotion Recognition.} For our compared baselines, EfficientFace \cite{CAER-Zhao} and MA-Net \cite{MA-Net} only considered facial features. Specifically, EfficientFace learned local and global face features by a local feature extractor and a channel-spatial modulator. MA-Net proposed to learn facial features by a multi-level attention mechanism. Other baselines were methods taking both facial and contextual features into consideration. CAER-Net \cite{CAER} and Jaiswal \textit{et al.} \cite{CAER-Jaiswal} both proposed to use spatial-wise attention mechanism to capture context. Zeng \textit{et al.} \cite{CAER-Zeng} designed a graph convolution network to capture semantic relationships among different small contextual regions. SIB-Net \cite{CAER-Li-IJCB} constructed a recurrent neural network to propagate the sequential influence in a face-body-scene order. Li \textit{et al.} \cite{CAER-Li-TAC} also explored regions that contribute more to the emotion, without explicitly model of the object-context relation. 

Comparison results are shown in Tab. \ref{Tab_caer2017_leaderboard}. As we can see, our method reaches a Top-1 Accuracy of 85.30\%/91.36\% under different backbone settings, which outperforms state-of-the-art baselines by 4.0\%/2.9\%, respectively. These results further demonstrate the effectiveness of our methods.

\subsection{Ablative Study}

In this section, we conduct ablative experiments to achieve a better understanding of the effect of each component in our Context-LGM. As shown in Tab. \ref{Tab_self_ablative}, the improvement compared to the vanilla cross-entropy method mainly comes from the latent generative model (VAE) that explicitly models the object-context correlation, as well as its following contextual posterior Transformer equipped with modified attention function to select contextual factors that are highly correlated with the object. These results indicate the importance of modeling the relation between contextual features and the object. 

\subsection{Comparison with Other Methods in Learning $z_c$} 

To further validate the superiority of incorporating object-context relation, we compare with other non object-context relation methods by replacing our Transformer with them when inferring $z_c$. These methods, as categorized based on their applications, include context blocks in neural architectures designs \cite{CVContext_Multi-Crop-17,CVContext_Non-Local-18,CVContext_SE-Net-18,CVContext_Stand-Alone-19,CVContext_LR-Net-19,CVContext_AA-Conv-19,CVContext_GloRe-19,CVContext_LCT-20,CVContext_ECA-Net-20}, image classification \cite{CVContext_SPP-Net-14,CVContext_Multi-Crop-17,CVContext_A2-Net-18,CVContext_SE-Net-18,CVContext_Stand-Alone-19,CVContext_LR-Net-19,CVContext_AA-Conv-19,CVContext_GloRe-19,CVContext_LCT-20,CVContext_ECA-Net-20}, detection \cite{CVContext_RCAM-21}, segmentation \cite{CVContext_DeepLab-v3plus-18,CVContext_Context-Enc-18,CVContext_Asy-Non-Local-19,CVContext_Att-Deeplab-v3-plus-20,CVContext_CaC-Net-20,CVContext_CG-Net-21}, scene paring \cite{CVContext_PSP-17,CVContext_DA-Net-19} and action recognition \cite{CVContext_GloRe-19}. Their corresponding context modeling methods contain stacking of different sizes feature maps \cite{CVContext_SPP-Net-14,CVContext_PSP-17,CVContext_Multi-Crop-17}, atrous convolution \cite{CVContext_DeepLab-v3plus-18}, spatial-wise attention \cite{CVContext_Non-Local-18,CVContext_Stand-Alone-19,CVContext_LR-Net-19,CVContext_AA-Conv-19}, channel-wise attention \cite{CVContext_Context-Enc-18,CVContext_SE-Net-18,CVContext_LCT-20,CVContext_ECA-Net-20}, spatial-channel wise attention \cite{CVContext_A2-Net-18,CVContext_DA-Net-19,CVContext_CaC-Net-20}, graph based spatial attention \cite{CVContext_GloRe-19}. Some of these works combined two of different types context modeling methods, such as \cite{CVContext_Asy-Non-Local-19} combined features pyramid with spatial-wise attention, \cite{CVContext_Att-Deeplab-v3-plus-20,CVContext_CG-Net-21,CVContext_RCAM-21} combined atrous convolution with channel-wise attention. All these methods only exploit object's label to supervise the learning of contextual feature; \emph{that is}, they overlook the importance of object-context relation.

As observed in Tab. \ref{Tab_cross_ablative}, our contextual posterior Transformer outperforms all the other methods by a noticeable margin (0.8\%-1.2\% in lung cancer prediction and 0.6\%-6.0\% in emotion recognition). These results show that, due to the diversity of contextual features, the existing methods may fail to capture them comprehensively; \textbf{in contrast}, the exploitation of an extra inductive bias, \textit{i.e.} object-context relation, enables our method to capture them well. 

\subsection{Hyper-parameters Analysis}

In this section, we examine the influence of hyper-parameters on our contextual posterior Transformer. 

\noindent \textbf{Layers Number.} The results are shown in Tab. \ref{Tab_hyper_paras_layernum}. It can be observed that our Transformer reaches satisfactory results when the layers number is $2$ for lung cancer prediction or $1$ for emotion recognition. As a matter of fact, since using more layers could significantly increase computation costs, similar layers number setting are also reported in some other visual Transformers, such as \cite{Trans_Meet_Track} use $1$ layer, \cite{Video_Act_Trans,SuperR_Trans} uses $3$ layers.

\noindent \textbf{Heads Number.} Multi-head mechanism is introduced in Transformer to model various aspects of contextual relationships. We conduct experiments on different heads number and report the results in Tab. \ref{Tab_hyper_paras_headnum}. We can see that our model reaches the best performance when $n_h=4$ for lung cancer prediction and $n_h=2$ for emotion recognition. This may be because contextual features in lung cancer are more complicated than those in emotion. As an intuitive example, in emotion recognition, there may be at most two types of contextual features (\textit{i.e.} gestures, other people) presenting together. However, in lung cancer, the malignant nodule can cause multiple pleural indentations (Fig. \ref{Fig_exp_cross_att_head}-a) or vascular convergence (Fig. \ref{Fig_exp_cross_att_all}-f) at the same time. 

\subsection{Visualization}

In Context-LGM, we use object-context correlation as extra information to locate contextual factors. Such a correlation is parameterized by the cross-attention block in our Transformer. This block computes a correlation matrix between object $z_c$ and surrounding regions in $F_{mask}$. Regions with high correlation values are contextual factors, while those with low correlation values are backgrounds.

\noindent \textbf{Learned Correlation Matrix.} We visualize the correlation metrics in Fig. \ref{Fig_exp_cross_att_all}. For each example, the left image is the input $x$ to our Context-LGM, and the right image shows the computed correlation matrix. In the correlation matrix, a large value indicates a high correlation with the object. The object regions are with zero response. This is because we mask them in $F_{mask}$ to avoid interference from object-object correlations.

We can observe that regions with contextual features show high correlation response, and those without context show low response. For example, in lung nodule related context, indentation to bottom right pleura in Fig. \ref{Fig_exp_cross_att_all}(a), and indentation to the left upper pleura in Fig. \ref{Fig_exp_cross_att_all}(b,c) are captured by high response correlation. In Fig. \ref{Fig_exp_cross_att_all}(d,e), the bronchial interruption context in the nodule's bottom areas is emphasized. In Fig. \ref{Fig_exp_cross_att_all}(f), the vascular convergence context is emphasized, too. Also, the pleural and vascular attachment context are well captured by object-context correlation in Fig. \ref{Fig_exp_cross_att_all}(g-i). In facial emotion related context, various of human gestures (Fig. \ref{Fig_exp_cross_att_all}-j,k,l,n,o,p,q) and posture (Fig. \ref{Fig_exp_cross_att_all}-m) are well captured by our object-context correlation. In Fig. \ref{Fig_exp_cross_att_all}(r-u), our method learns correlation between the target face and other people. In Fig. \ref{Fig_exp_cross_att_all}(v,w), it can be seen that our object-context correlation is also able to capture the clothing context (business suit with serious facials, messy collar with sad facials). Also, context from environmental objects such as phones and gifts are well capture in Fig. \ref{Fig_exp_cross_att_all}(x,y).

From these observations, we could conclude that our object-context correlation mainly looks at contextual regions in an object's surroundings, thus well promote the learning of contextual representations.

\noindent \textbf{Comparison with Other Methods.} We also compare with context captured by other methods and show the results in  Fig.~\ref{Fig_exp_visu_compare}. As we can see, only our Context-LGM correctly capture the pleural indentation and bronchial interruption context in Fig.~\ref{Fig_exp_visu_compare}(a) and the bronchial interruption context in Fig.~\ref{Fig_exp_visu_compare}(b). In Fig.~\ref{Fig_exp_visu_compare}(c), only our method comprehensively captures both the gesture and other people context. Such a more accurate and comprehensive capturing of context further verify the effectiveness of using object-context relation, as well as the use of multi-head attention as implementation.

\noindent \textbf{Roles of Different Heads.} Each of our cross-attention block composites multiple attention heads ($n_h=4$ in lung cancer prediction and $n_h=2$ in emotion recognition, specifically). This mechanism could enhance representation powers on different aspects of contextual information. To show the roles of different attention heads and how do they work together, we present visualization examples of correlation metrics in different attention heads in Fig. \ref{Fig_exp_cross_att_head}

It can be observed that different heads seem to focus on different regions. For example, nodule in Fig. \ref{Fig_exp_cross_att_head}(a) shows three indentations to its bottom right pleura. It seems that head $1$ mainly looks at the left indentation, head $3$ focuses on the middle indentation, and heads $2,4$ mainly correlate with the right indentation. In Fig. \ref{Fig_exp_cross_att_head}(b), the nodule shows both pleural indentation context in the upper left and bronchial interruption context in the right bottom. It seems that heads $1,4$ focus on correlation with the first context regions, and heads $2,3$ focus on correlation with the second context regions. In Fig. \ref{Fig_exp_cross_att_head}(c), contextual factors include both the target women's gesture and other people. It seems that head 1 mainly looks at the other people context, and head 2 focuses on the gesture context.

Such an observation indicates complementary cooperation may be learned among different attention heads. This mechanism could be especially beneficial when multiple contextual regions exist.

\section{Conclusion}
\label{Sect_conclusion}

In this paper, we propose Context-LGM, a novel latent generative model for context-aware object recognition. We firstly incorporate the object-context relation via their generating processes in the latent generative model. Then, we design a reformulated VAE framework, within which a carefully modified Transformer is equipped to take the object's information as a reference and infer contextual features.
Our method can achieve state-of-the-art results on lung nodule prediction and emotion recognition. Moreover, the learned context features are highly explainable. 

We believe that the object-context relation, as widely exists in other vision tasks such as scene understanding, action recognition, and human-computer interaction, can serve a broader family of applications. We leave this exploration in the future work.


%

\appendices
\section{Pathological Background} 
\label{Sect_appendix}
Though context in the natural image is easy to understand, it takes some background knowledge to understand it in the medical image. To give an intuition, we provide radiological examples about the common context in lung cancer prediction.

\begin{figure}[h]
\begin{minipage}[d]{0.1\linewidth}
\centering
\includegraphics[width=3.0in]{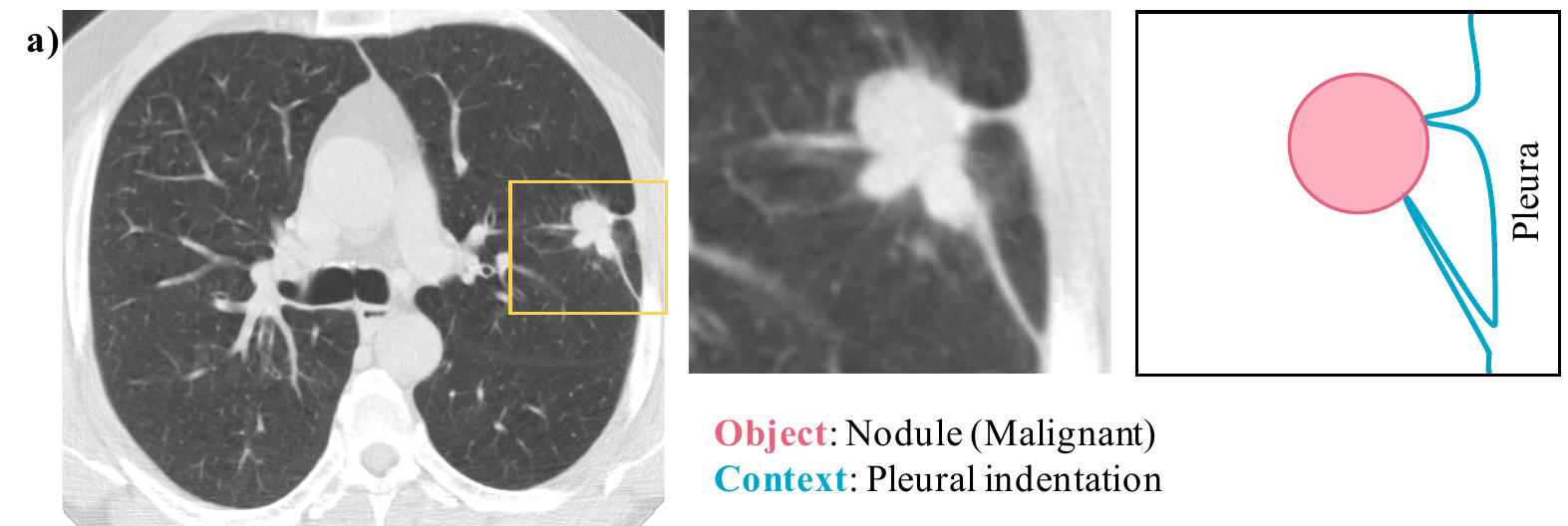}\\
\vspace{0.1cm}
\includegraphics[width=3.0in]{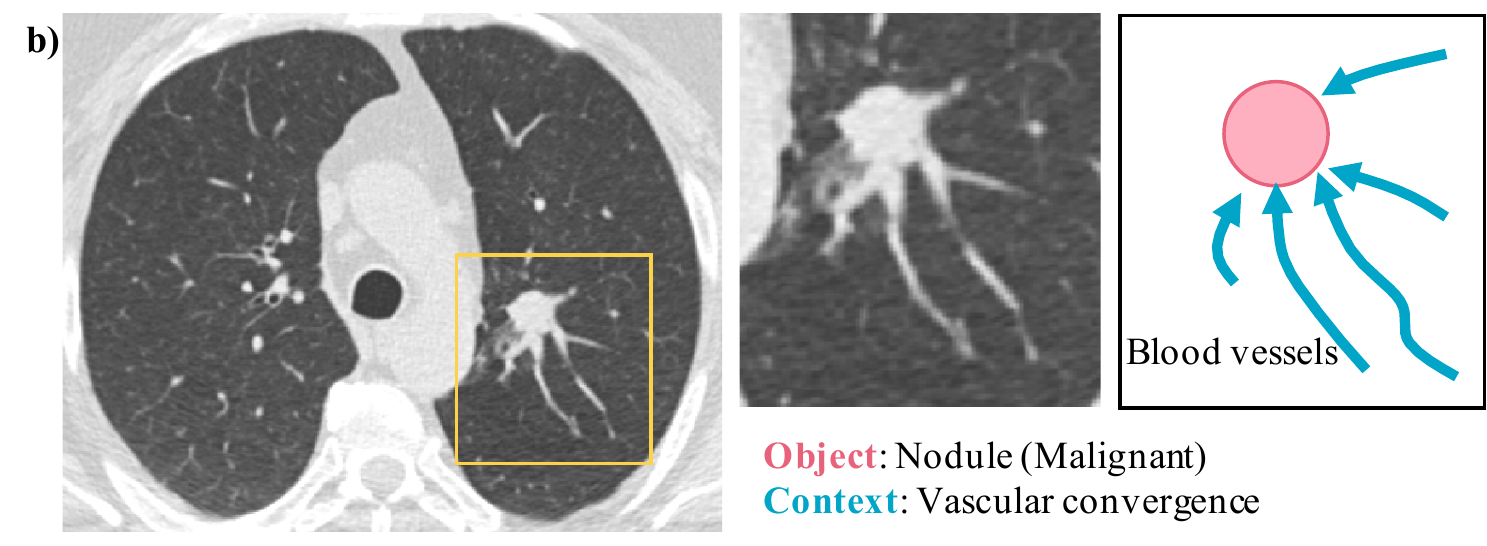}\\
\vspace{0.1cm}
\includegraphics[width=3.0in]{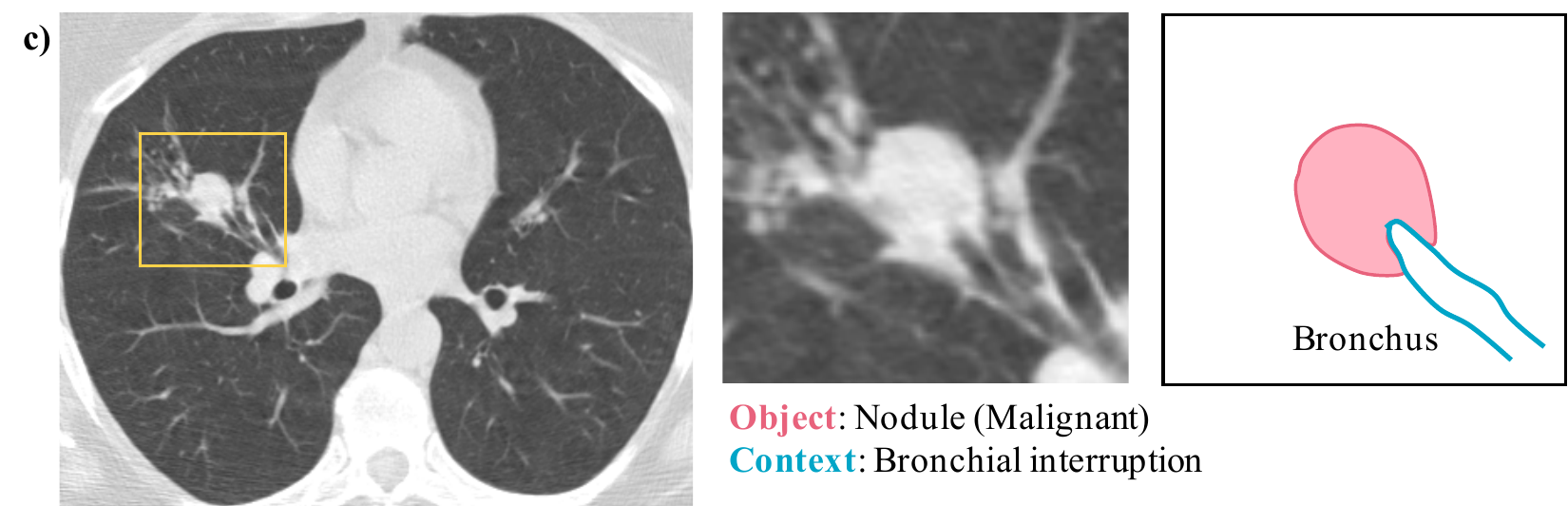}\\
\vspace{0.1cm}
\includegraphics[width=3.0in]{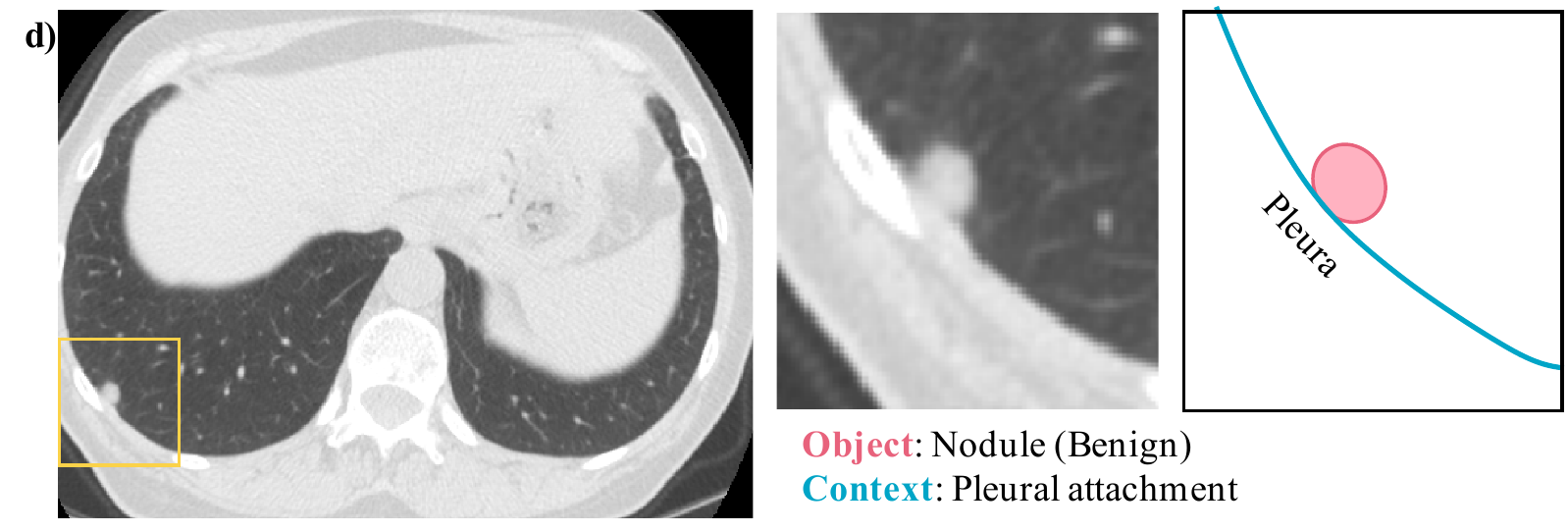}\\
\vspace{0.1cm}
\includegraphics[width=3.0in]{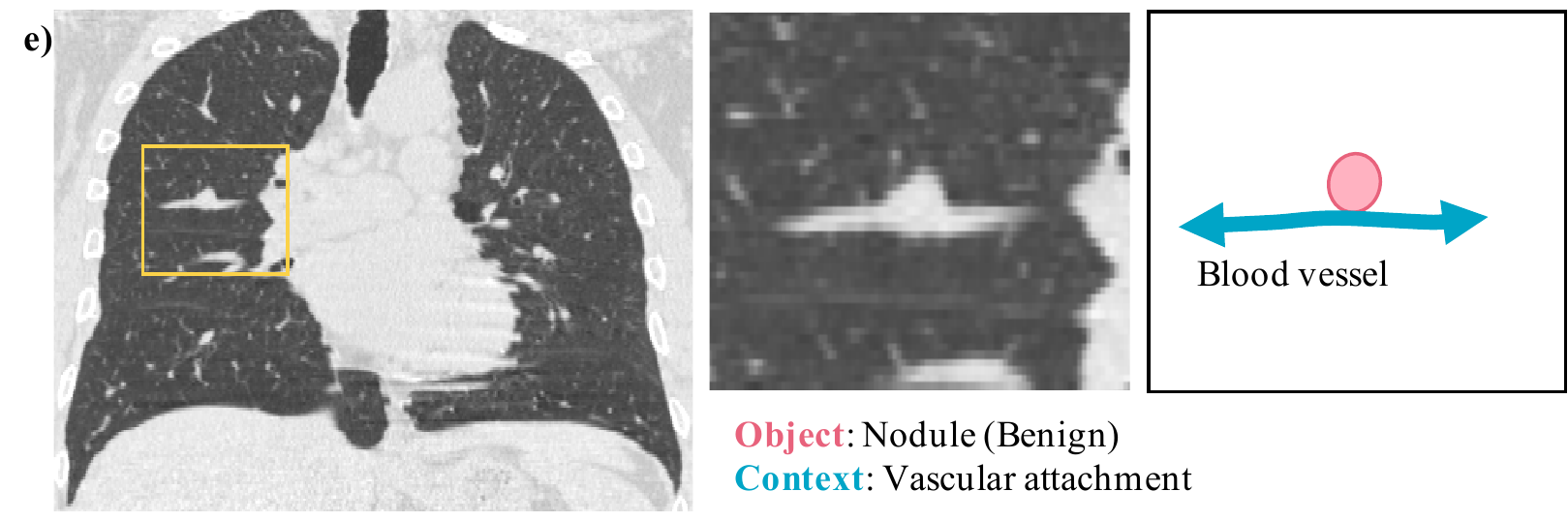}\\
\vspace{0.1cm}
\end{minipage}%
\caption{Radiological examples for context in lung cancer prediction.}
\label{Fig_pathological_background1}
\end{figure}

In Fig. \ref{Fig_pathological_background1}, (a-c) show context correlated with malignant nodules, and (d-e) show context correlated with benign nodules.

Specifically, \textbf{pleural indentation} \cite{Pathology_PleuralIndentation} in Fig. \ref{Fig_pathological_background1}(a) represents pulling the visceral pleura towards the nodule. It suggests a possible pleural invasion by peripheral lung tumor. Invasion and contractile changes of the tumor cause dead space, then the linear strand between tumor and pleural is formed by compensatory expansion of lung tissues to fill the dead space; \textbf{vascular convergence} \cite{Pathology_VasConverge} in Fig. \ref{Fig_pathological_background1}(b) is described as vessels converging towards a nodule. This is because angiogenesis is essential for tumor growth and metastasis. The vascular endothelial growth factor is then synthesized continuously and excessively in the tumor, promoting the proliferation of vascular endothelial cells during vessel formation; \textbf{bronchial interruption} \cite{Pathology_Bronchi} in Fig. \ref{Fig_pathological_background1}(c) means the bronchus is obstructed abruptly by the tumor or it penetrates into the tumor with tapered narrowing and interruption. This context is resulted by the growth pattern of malignant nodules, i.e. the hilic growth and the lepidic growth. In hilic growth, the tumor cells proliferate and pile up continuously, forming a solid mass and obstruct adjacent bronchus. In lepidic growth, the tumor cells line the alveolar wall and directly spread from one alveolus to another through the pore of K\"ohn. Thus the bronchus remains intact and penetrates into the mass. \textbf{Pleural attachment} in Fig. \ref{Fig_pathological_background1}(d), and \textbf{vascular attachment} in Fig. \ref{Fig_pathological_background1}(e) respectively represent that the nodule is adjacent to a pleura or blood vessel. Their relation to benignity are mainly confirmed by clinical statistics \cite{Pathology_Attachments}.

\section*{Acknowledgment}

The work is supported in part by NSFC Grants (No. 61625201), MOST Grants (No. 2018AAA0102004), and the Beijing Municipal Science and Technology Planning Project (No. Z201100005620008).
\ifCLASSOPTIONcaptionsoff
  \newpage
\fi



\bibliographystyle{IEEEtran}
\bibliography{IEEEabrv,reference}
\end{document}